\documentclass{article}

    \usepackage[preprint]{neurips_data_2024}

\usepackage[utf8]{inputenc} 
\usepackage[T1]{fontenc}    
\usepackage{url}            
\usepackage{booktabs}       
\usepackage{amsfonts}       
\usepackage{nicefrac}       
\usepackage{microtype}      
\usepackage{multirow}
\usepackage{amsmath}
\usepackage{wrapfig}
\usepackage{framed}
\usepackage[table,dvipsnames]{xcolor} 
\usepackage[colorlinks=true,linkcolor=MidnightBlue,anchorcolor=black,citecolor=MidnightBlue,urlcolor=MidnightBlue, pdfborderstyle={/S/U/W 0.5}]{hyperref} 

\newcommand{\CC}{\cellcolor{gray!13}}

\usepackage[framemethod=TikZ]{mdframed}
\mdfdefinestyle{PVFrame}{%
    frametitlerule=true,
    nobreak=true,
    frametitlebackgroundcolor=black,
    frametitlefont={\bfseries\color{white}},
    linecolor=black,
    outerlinewidth=0pt,
    roundcorner=0pt,
    innertopmargin=4pt,
    innerbottommargin=4pt,
    innerrightmargin=8pt,
    innerleftmargin=8pt,
    backgroundcolor=black!5}

\title{$\forall$uto$\exists$val: Autonomous Assessment of LLMs in\\Formal Synthesis and Interpretation Tasks}

\author{%
    Rushang Karia$^*$, Daniel Bramblett$^*$, Daksh Dobhal, Pulkit Verma, {\normalfont and}  Siddharth Srivastava \\
    School of Computing and Augmented Intelligence \\
    Arizona State University\\
    Tempe, AZ 85281 \\
    \texttt{\{rushang.karia,drbrambl,ddobhal,verma.pulkit,siddharths\}@asu.edu} \\
}

\usepackage{amsthm}
\theoremstyle{definition}
\newtheorem{definition}{Definition}[section]

\newcommand{\FS}{\textit{FS}}
\newcommand{\NL}{\textit{NL}}
\newcommand{\LLM}{\ifmmode L_{|\theta} \else \(L_{|\theta}\)\fi}
\newcommand{\FSNL}{\ifmmode\FS\rightarrow\NL\else\(\FS\rightarrow\NL\)\fi}
\newcommand{\NLFS}{\ifmmode\NL\rightarrow\FS\else\(\NL\rightarrow\FS\)\fi}
\newcommand{\FTT}{\ifmmode\NL\leftrightarrow\FS\else\(\NL\leftrightarrow\FS\)\fi}
\newcommand{\APPROACH}{$\forall$uto$\exists$val}

\usepackage{graphicx}
\usepackage{algorithm} 
\usepackage{algpseudocode}

\usepackage{qtree}

\begin{document}

\maketitle
\def\thefootnote{*}\footnotetext{These authors contributed equally to this work}\def\thefootnote{\arabic{footnote}}

\begin{abstract}
This paper presents $\forall$uto$\exists$val, a new approach for scaling LLM assessment in translating formal syntax -- such as first-order logic, regular expressions, etc -- to natural language (interpretation) or vice versa (compilation), thereby facilitating their use in applications such as generating/explaining logic and control flow for programs etc. Existing approaches for LLM assessment in these areas require labor-intensive ground-truth creation, the availability of which undermines the separation of training and test sets. Furthermore, such datasets typically include relatively few hand-coded test cases over which LLM accuracy is determined, thus making them inadequate for determining the safety or correctness of their generated outputs. We introduce a new approach that utilizes context-free grammars (CFGs) to generate out-of-distribution datasets on the fly and perform closed-loop testing of LLM capabilities using formal verifiers to guarantee the correctness of LLM outputs without any human intervention. We release our dataset and benchmark as open-source code at \url{https://github.com/AAIR-lab/auto-llm-assessment}. We also conduct an assessment of several SOTA closed and open-source LLMs to showcase the feasibility and scalability of this paradigm. Our experiments reveal that SOTA LLMs are unable to solve the formal translation task adequately.
\end{abstract}

\section{Introduction}

Foundation Models such as Large Language Models (LLMs) have been demonstrated to successfully perform many natural language tasks such as translation with human-like performance \citep{DBLP:conf/naacl/XueCRKASBR21}. Furthermore, LLMs have been shown to be versatile and easily transferrable to several downstream, domain-specific tasks (e.g., French-to-English Translation \citep{zhu2023multilingual}) using approaches such as supervised fine tuning (SFT) \citep{lu2023instag,zeng2024teaching}. 
Recently, there has been research on using LLMs to translate natural language (\NL{}) to formal syntax (\FS{}) for robotics \citep{liang2023code} and for constructing planning problems solvable using existing solvers \citep{guan2023leveraging}. Although these methods have been successful in small-scale scenarios, their effectiveness in accurately translating \NL{} to/from \FS{} remains uncertain. This uncertainty mainly stems from the difficulty in assessing how good LLMs are for truth maintenance in such tasks.

This paper addresses three key questions for performing this evaluation: \emph{(1) How do we accurately assess an LLM's translation abilities? (2) Can we avoid relying on datasets that require human-annotators? (3) Can we construct dynamic datasets that can be scaled as LLMs are retrained?}
Existing methods for LLM assessment are limited along one or more of these dimensions. Firstly, the datasets used are static, making them prone to be overfitted/memorized as they are likely included in the training sets when LLMs are retrained (e.g., \citep{han2022folio,sinha2019clutrr}). Secondly, these datasets are not easy to scale since they need to be annotated by humans. Finally, these evaluation methodologies only check a few handwritten golden examples thus assessing a limited set of input parameter configurations for the formal syntax (e.g., \citep{chen2021evaluating}). 

\textbf{Main contributions} This paper introduces a new approach for autonomous assessment of LLMs in a scalable fashion. Our key contributions are as follows:
\begin{enumerate}
    \item A new approach for automatic synthesis of scalable evaluation datasets 
    that are likely to be unseen in the LLM's training set.
    \item The utilization of formal verifiers such as theorem provers to provably validate syntax-independent notions of correctness without having to exhaustively test over all possible truth valuations of formal syntax involving logic.
    \item \APPROACH{}: a scalable, plug-and-play assessment system for new LLMs as they are developed.
    \item  Extensive empirical evaluation of SOTA LLM capabilities in three popular forms of formal syntax: propositional logic, first-order logic, and regular expressions. Our experiments reveal the strengths of our approach, the limitations of LLMs, and encourage new research in the area.
\end{enumerate}

This paper is organized as follows: The next section provides the necessary formal framework and defines the formal syntax translation task. Sec.\,\ref{sec:nlfs} describes our approach for generating datasets and conducting autonomous assessment of LLM-generated outputs. Sec.\,\ref{sec:datasets} provides details about the datasets. We then present an evaluation of SOTA LLMs using our approach (Sec.\,\ref{sec:assessment}). Next, we provide an account of related work in the area (Sec.\,\ref{sec:related_work}). Finally, we state our conclusions, acknowledge some of the limitations of our current work, and discuss avenues for future work in Sec.\,\ref{sec:conclusions}. 

\section{Formal Framework}
\label{sec:formal_framework}

\textbf{(Large) Language Models: (L)LMs} LMs are non-linear functions represented by a set of parameters $\theta$ that, given a set of input tokens $x$, typically representing \NL{}, predict the output token $y$ using the distribution $P(y|x;\theta)$. The input tokens consist of a \emph{prompt} that provides instructions about the task. It has been shown that correctly engineering the prompts can significantly alter the quality of the predictions \citep{sahoo2024systematic}. LLMs are LMs with billions of parameters.

\textbf{Propositional Logic} Also known as propositional calculus, propositional logic is a branch of logic that utilizes \emph{propositions} and \emph{logical operators} to construct arguments for reasoning. Propositions are represented using \emph{variables} and evaluate to either true or false. These variables can be associated with \NL{} sentences. Propositions can be combined using logical operators such as negation ($\neg$), conjunction ($\land$), and disjunction ($\lor$) to form compound \emph{sentences} $P$\footnote{Implications $(\Rightarrow)$ and Bi-conditionals $(\Leftrightarrow)$ can be expressed using $\land, \lor$, and $\neg$}. These sentences can then be used to perform reasoning using the rules of logic. For example, propositions, $p_1 = \textit{It is raining}$, $p_2 = \textit{It is sunny}$ can be used to create a sentence $P = p_1 \lor p_2$. If $P$ is true and $\neg p1$ is observed, then one can use the rules of inference to deduce that $p_2$ must be true \citep{DBLP:books/daglib/0017977}.

\textit{Equivalence in Propositional Logic} Two sentences in propositional logic, $P_1$ and $P_2$, are equivalent, $P_1 \equiv P_2$, iff their truth values agree for all possible assignments. E.g., $\neg (p_1 \land p_2) \equiv \neg p_1 \lor \neg p_2$.

\textbf{First-order Logic} First-order logic differs from propositional logic in that sentences are constructed using \emph{predicates} and \emph{objects}. Predicates are used to represent relations between objects, e.g. $\textit{Married}(\textit{John}, \textit{Alice})$. Furthermore, it introduces the use of \emph{quantified variables} using the universal ($\forall$) and existential ($\exists$) operators to allow reasoning over sets of objects. A 
popular example is the syllogism where given two first-order logic sentences, $\forall x. \text{ }\textit{Man}(x) \rightarrow \textit{Mortal}(x)$, $\textit{Man}(\textit{Socrates})$, one can conclude that $\textit{Mortal}(\textit{Socrates})$. A first-order logic sentence $F$ can be interpreted using a universe $\mathcal{U}$, a substitution operator $\sigma$, and an interpretation function $\mathcal{I}$ \citep{DBLP:books/aw/RN2020}.

\textit{Equivalence in First-order Logic} Two sentences, $F_1, F_2$ in first-order logic are equivalent $F_1 \equiv F_2$ iff they are equivalent under all possible interpretations. E.g., $\neg \forall x. \text{ } \textit{pred}_1(x) \equiv \exists y. \text{ } \neg \textit{pred}_1(y)$.

\textbf{Regular Expressions} A regular expression (regex) is a sequence of characters that can be used to determine whether a particular string matches the pattern or \emph{language} induced by the regex. A regex is defined over an alphabet $\Sigma \cup \{\varepsilon\}$ where $\varepsilon$ is the empty string. One can then construct regexes using $\Sigma$ and operators like the Kleene star $\ast$ which represents zero or more occurrences of a given pattern. For example, the regex $2(01)^\ast$2 using $\Sigma = \{0, 1, 2\}$ matches all strings possible using $\Sigma$ that begin with a two, contain zero or more instances of alternating zeroes, and end with a two. Regular expressions can be converted to Deterministic Finite Automata (DFAs) which are finite-state machines that can easily determine whether a given string matches the regex. Furthermore, each regex can be converted into a minimal DFA, unique up to isomorphism \citep{hopcroft2001introduction}.

\textit{Equivalence between Regular Expressions} Two regular expressions, $R_1$ and $R_2$ are equivalent $R_1 \equiv R_2$ if they represent the same language. One way to check this is to reduce both $R_1, R_2$ to their corresponding minimal DFAs $D_1, D_2$ representing $R_1, R_2$ respectively. It is known that $R_1 \equiv R_2$ if the corresponding minimal DFA's are isomorphic, i.e, $D_1 \simeq D_2$ \citep{hopcroft2001introduction}. 

Propositional logic, first-order logic, and regexes are of profound significance and have several real-world applications. Propositional logic is used for the automatic synthesis and verification of systems \citep{sat_synthesis,sat_verification}. First-order logic is used by automated planning systems for performing reasoning \citep{DBLP:journals/tissec/HalpernW08}, for writing formal safety specifications \citep{DBLP:journals/puc/RanganathanC03,DBLP:conf/vmcai/BloemKS14}. Similarly, regular expressions and DFAs are of widely used in describing the control flow of programs \citep{DBLP:journals/smr/MorrisGF97}.

We refer to sentences (strings) in first-order and propositional (regexs) as formal syntax \FS{} in this paper. We now provide a definition of \textit{Interpretation} and \textit{Compilation} in the context of LLMs and \FS{}.

\begin{definition}[Interpretation: \FSNL] 
Given an LLM \LLM{} and an expression $\varphi$ in \FS{}, then the interpretation is defined as using \LLM{} to convert $\varphi$ to an \NL{} description $\Phi$.
\end{definition}

\begin{definition}[Compilation: \NLFS]
Given an LLM \LLM{} and an \NL{} description $\Phi$, then compilation is defined as using \LLM{} to convert $\Phi$ to \FS{} $\varphi$.
\end{definition}

We refer to interpretation and compilation as ``formal translation'' (\FTT{}) in this paper. Thus, \FTT{} represents an LLM's capability to perform formal syntax translation in either direction and the assessment task is known as LLM assessment of \FTT{}. Naturally, LLMs may not compile or interpret accurately due to issues such as hallucination \citep{Ji_2023}, etc. 

\section{Our Approach: \APPROACH{} for Autonomous Assessment of \FTT{}}
\label{sec:nlfs}

\begin{figure}[t]
    \centering
    \includegraphics[width=\linewidth]{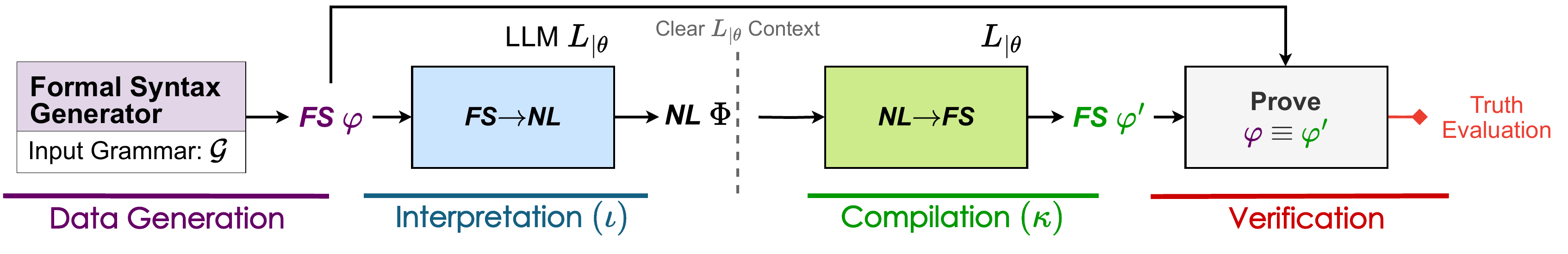}
    \caption{Our overall process for autonomous assessment of \FTT{}.}
    \label{fig:fig1}
    \vspace{-0.15in}
\end{figure}

We now describe our approach, \APPROACH{}, for assessing \FTT{} without any human intervention. Our approach (a) provides dynamically generated datasets that can be scaled arbitrarily using context-free grammars to generate datasets with a large amount of out-of-distribution ground-truth data, (b) uses intrinsic LLM capabilities to automatically assess \FTT{} without requiring any labeled annotations of \NLFS{} or \FSNL{}, and (c) uses formal verifiers to rigorously check and guarantee correctness of \FTT{} without having to exhaustively iterate over all possible interpretations. A description of our dataset generation techniques, our pipeline for \FTT{}, and our usage of verifiers for assessing \FTT{} is provided below.

\textbf{Context-Free Grammars (CFGs)} CFGs are formal grammars that consist of \emph{production rules} over \emph{terminal} and \emph{non-terminal} symbols. These rules can be applied to generate sentences. The grammars are context-free since any of the production rules for a non-terminal symbol can be applied regardless of context. CFGs have important real-life applications. For example, in computing, CFGs are used to describe the structure of source code in programming languages, and in linguistics, to describe the structure of sentences \citep{DBLP:books/lib/JurafskyM09}. 

An example CFG $\mathcal{G}$ with 4 production rules is as follows: (i) $S \rightarrow AB$, (ii) $A \rightarrow aA$, (iii) $A \rightarrow \varepsilon$, (iv) $B \rightarrow b$.  
$\mathcal{G}$ consists of a set of non-terminals $S, A, B$, a set of terminals $a, b, \varepsilon$, (since there are no production rules where they appear on the LHS). $\varepsilon$ is the \textit{empty string}. A non-terminal that has no other ancestors other than itself is a \emph{start symbol} ($S$ in this example) and is used to create strings by application of the production rules. $\mathcal{G}$ can be used to generate all possible strings of the form $a^*b$, i.e., strings containing one or more $a$ characters followed by a $b$. This is achieved by repeatedly applying production rules of $\mathcal{G}$ resulting in an \emph{infix parse tree} that encodes the order of the rules applied. The depth of this tree is often used to measure the \emph{descriptional complexity} of a given string generated using the CFG. \citep{DBLP:journals/tcs/Csuhaj-VarjuK93}.

We use CFGs to dynamically generate our datasets. Thus, our approach can easily scale up the size of datasets. Another advantage is that the grammars can be tweaked to generate datasets whose ground-truth data possesses specific properties. For example, a dynamic dataset that only consists of $k$--SAT sentences -- propositional logic in the Canonical Normal Form
$(P_i \lor \ldots \lor P_k) \land (P'_i \lor \ldots \lor P'_k) \land \ldots$ where $P_i \in \{ p_j, \neg p_j\}$ -- can be easily generated thus allowing for scalable assessment in specific downstream tasks, e.g., \textit{Can LLMs compile \NL{} as $k$-SAT formulae or interpret $k$-SAT as \NL{}?}.

\textbf{Automatic Formal Translation} Traditionally, most methods only focus on one direction of assessment, \NLFS{} or \FSNL{} but not both since hand-annotated ground truth data would be required for the {\color{red} intermediate} step ($\FS{} \rightarrow {\color{red}\NL{}} \rightarrow \FS{}$ or $\NL{} \rightarrow {\color{red}\FS{}} \rightarrow \NL{}$) as well. Our approach automatically assesses both directions simultaneously.

Our approach is based on the following intuition. Let $\iota$ be a non-deterministic function that maps \FS{} to \NL{}. Similarly, let $\kappa$ be a non-deterministic function that maps \NL{} to \FS{}. $\iota$ ($\kappa$) then serve as an interpreter (compiler) that can perform \FSNL{} (\NLFS{}) respectively. In general, there are many possible correct interpretations of $\varphi \in \FS{}$ and many correct compilations of $\Phi \in \NL{}$.  Thus, $\iota$ and $\kappa$ are not injective (1:1) functions and thus $\iota^{-1}$ and $\kappa^{-1}$ are not well-defined.

Our key observation is that if $\iota$ and $\kappa$ come from the same system (e.g., a neural network or LLM), then we can check the accuracy of the system by \emph{composing} $\iota$ and $\kappa$. Let $\varphi$ be any arbitrary \FS{} expression and let \LLM{} be an LLM. Now, if \LLM{} preserves truth in both translations, then $\Phi = \iota_{\LLM{}}(\varphi)$ will be a factual \NL{} interpretation of $\varphi$ and $\varphi' = \kappa_{\LLM{}}(\Phi)$ will be a factual \FS{} compilation of $\Phi$. Since, $\Phi$ is a natural language description, it is quite challenging to check whether $\iota_{\LLM{}}(\varphi)$ is a factual interpretation of $\varphi$ without human intervention. However, since the same LLM is capable of both interpretation and compilation, $\varphi' = \kappa(\iota(\varphi))$ will be a factual representation of $\varphi$ even if they are not syntactically identical. Thus, we only need to check if $\varphi \equiv \varphi'$.\\
\textit{Example} Let $\varphi = p_1 \land p_1$,
$\Phi = \iota_{\LLM{}}(\varphi) = $ \textit{A conjunction of propositions $p_1$ and $p_1$}, and $\varphi' = \kappa_{\LLM{}}(\Phi) = p_1$. It is very difficult to check if $\Phi$ is a factual representation of $\varphi$, but easy to check if $\varphi \equiv \varphi'$ using a formal verifier.

\textbf{Formal Verification} Since \APPROACH{} uses formal syntax $\varphi$ as input and produces formal syntax $\varphi'$ as output, we can use formal verifiers to check whether $\varphi \equiv \varphi'$. Key advantages of this approach is that it guarantees correctness over all possible inputs. Due to its practical significance, formal verification has been used in a large variety of systems \citep{DBLP:journals/todaes/KernG99}. Recently, this approach has also been used in conjunction with LLMs for logical reasoning \citep{DBLP:conf/nips/YangSGCSYGPA23} but its application in \FTT{} has, to the best of our knowledge, not been explored.

We use the above insights to automatically assess $\iota = \FSNL{}$ and $\kappa = \NLFS{}$ capabilities of LLMs by using the same LLM \LLM{} to represent $\iota$ and $\kappa$. Since LLMs utilize context windows that include the input $x$ and all output $y_0, \ldots, y_i$ generated so far to predict the next token $y_{i+1} \sim P(\cdot|x, y_0, \ldots, y_i; \theta)$, we clear the context of \LLM{} after \FSNL{} to ensure that no contextual knowledge is exchanged between the interpretation and compilation process.

Fig.\,\ref{fig:fig1} shows our overall assessment process. We use a CFG $\mathcal{G}$ to automatically generate a dataset $\mathcal{D}$ consisting of elements of \FS{}. Next, using \FS{}  $\varphi \in \mathcal{D}$, we use an LLM \LLM{} to interpret $\varphi$ as \NL{} $\Phi$ using a prompt designed for \FSNL. Next,  the context of \LLM{} is cleared, and the \NL{} description $\Phi$ is compiled to \FS{} $\varphi'$ using a prompt designed for \NLFS{}. We then use a verifier (e.g., Z3 \citep{DBLP:conf/tacas/MouraB08}, Prover9 \citep{prover9-mace4}) to assess if $\varphi \equiv \varphi'$ since both are elements of \FS{}.

\section{Datasets and Assessment Framework}
\label{sec:datasets}

Our goal is to provide an automatic framework for researchers to assess LLM capabilities for formal translation. As a result, we have designed a convenient framework with packaged datasets and an automatic approach for evaluating pass@k scores \citep{chen2021evaluating}. Our framework is written in Python 3, includes several pre-computed datasets to conduct evaluations, and allows researchers to easily modify them, add new datasets, prompts, LLMs, etc. We now describe the datasets and benchmarks that any newly developed LLM can be evaluated using \APPROACH{} out-of-the-box.

\subsection{Dynamic Dataset Generator and Pre-generated Datasets}
\label{subsec:dataset_desc}

\begin{figure}[!htbp]
\vspace*{-18pt}
\begin{minipage}[b]{0.20\linewidth}
    \begin{align*}
            S &\rightarrow S \land S \\
            S &\rightarrow (P \lor P \lor P) \\
            P &\rightarrow \neg v | v
    \end{align*}
    \begin{center}
        $(a)$ $3$--SAT
    \end{center}
\end{minipage}
\begin{minipage}[b]{0.25\linewidth}
    \begin{align*}
            S &\rightarrow (S \land S) \\
            S &\rightarrow (S \lor S) \\
            S &\rightarrow (\neg S) \\
            S &\rightarrow \neg v | v
    \end{align*}
    \begin{center}
        $(b)$ Propositional Logic
    \end{center}
\end{minipage}
\begin{minipage}[b]{0.27\linewidth}
    \begin{align*}
            S &\rightarrow Q \\
            Q &\rightarrow F|(\forall f. \text{ } Q) | (\exists f. \text{ } Q)\\
            F &\rightarrow (F \land F) | (F \lor F) \\
            F &\rightarrow (\neg F) |  \neg p | p
    \end{align*}
    \begin{center}
        $(c)$ First-order Logic
    \end{center}
\end{minipage}
\begin{minipage}[b]{0.25\linewidth}
    \begin{align*}
            S &\rightarrow (S)K \\
            S &\rightarrow S \Sigma K\\
            S &\rightarrow \Sigma K\\
            K &\rightarrow \ast | \varepsilon
    \end{align*}
    \begin{center}
        $(d)$ Regular Expression
    \end{center}
\end{minipage}
\caption{Grammars (described in Sec.\,\ref{subsec:dataset_desc}) used for synthesizing the datasets in \APPROACH{}.}
\label{fig:fig2}
\end{figure}

We provide 5 datasets of size $|\mathcal{D}|$  $\sim 20k$ samples each and constructed using our dataset generator. Each dataset corresponds to one of the grammars shown in Fig.\,\ref{fig:fig2} along with an additional dataset for first-order logic (described below). Each dataset has been partitioned into 10 batches of $\sim 2k$ samples each.  
All datasets are generated based on the descriptional complexity, however other metrics (e.g., total number of propositions, \# of states of minimal DFA etc.) for categorization are also available. Each dataset has roughly 500 samples per category for a convenient way to assess how LLM performance changes as a factor of the descriptional complexity. These divisions allow for easy computation of statistical measures (e.g. averages, standard deviations, etc) by using a single $2k$ batch or any subset of the batches. We have also extensively engineered prompts for these datasets to allow for comparing LLM capabilities without investing effort in prompt engineering. Our overall dataset's total number of unique samples is $|\mathcal{D}^\ast| $ is $\sim 85k$. We describe them below.

\textbf{$3$--SAT$(n)$ Dataset} ($|\mathcal{D}^\ast| \sim 10k$) This dataset uses the grammar from Fig.\,\ref{fig:fig2}a. The terminal $v$ is replaced with a vocabulary of $n$ propositions $p_1, \ldots, p_n$. This dataset is useful for calibrating the prompts since the sentences are highly structured (discussed in Sec.\,\ref{subsec:prompts}).

\textbf{Propositional Logic$(n)$ Dataset} ($|\mathcal{D}^\ast| \sim 19k$) Fig.\,\ref{fig:fig2}b indicates the grammar used for this dataset. Similar to $k$--SAT, this dataset also replaces terminals by randomly selecting from a list $n$ propositions.

\textbf{First-order Logic$(n_p, n_o)$ Synthetic and English Vocabulary Datasets} ($|\mathcal{D}^\ast| \sim 19k$ each) Fig.\,\ref{fig:fig2}c generates first-order logic sentences in the Prenex Normal Form where all quantifiers appear first. The terminals $p$ are replaced with predicates of the form $p(v_1, \ldots, v_n)$ where $p_i$ is a predicate name selected from a list of $n_p$ predicates, $v_i$ is either an object $o$ from a list of $n_o$ objects or is a free variable $f \in \{x_1, x_2, \ldots\}$ that is appropriately annotated within the scoping rules of the parentheses.\footnote{We use CFGs to generate structured sentences and allow hooks to add context-specific information}  The objects and predicate names are auto-generated synthetically for the synthetic version of the dataset. The English-version of the dataset uses VerbNet \citep{schuler2005verbnet} for predicate names, and Faker \citep{Faraglia_Faker} for object names.

\textbf{Regular Expression$(n)$ Dataset} ($|\mathcal{D}^\ast| \sim 18k$) Fig.\,\ref{fig:fig2}d specifies the grammar provided for generating regular expressions. The vocabulary $\Sigma$ is the set $\{0, \ldots, n-1\}$ where $n$ is a user-specified constant.

\emph{Fast Dataset Generation} Since CFG parse trees can quickly grow in size, we provide a way to perform a random walk of the grammar to quickly generate data. This ensures that our dataset can be scaled to generate expressions at larger depths without imposing significant computational requirements or constraints. Details of our implementation are included in the supplementary material.

\emph{Robust Parser} The grammars above generate strings that are parsed into expressions by the Natural Language Toolkit (NLTK) library \citep{DBLP:books/daglib/0022921} and stored in the dataset. The NLTK library performs several simplifications that prevent noise (e.g., correctly removing unnecessary parentheses, etc.) Similarly, we also use these libraries to parse the LLM outputs, thus providing a robust handling of any errors that can arise. Similarly, regexes are parsed using Reg2Dfa \citep{Reg2Dfa}. Any LLM output that cannot be parsed using these libraries is said to be \emph{syntactically non-compliant}. 

\subsection{Assessment Metrics}
\label{subsec:metrics}
\APPROACH{} automatically assesses LLMs and provides reports that help answer the questions below.

\textbf{A1. Are LLMs syntactically compliant when performing \FTT{}?} \APPROACH{} reports the syntactic compliance of LLMs as the total number of \FS{} expressions that could be successfully parsed. This allows to check if LLMs are able to successfully generate syntactically correct outputs. 

\textbf{A2. Are LLMs able to perform \FTT{}?} We compute the accuracy of the LLMs w.r.t. \FTT{}. A high accuracy indicates that LLMs are able to do both \FSNL{} and \NLFS{} well.

\textbf{A3. Can LLMs be used as verifiers?} \APPROACH{} computes the precision, sensitivity, specificity, $F_1$ scores, etc., using standard notations of the confusion matrix \citep{DBLP:journals/prl/Fawcett06} by utilizing LLMs to ask if $\varphi \equiv \varphi'$ for two \FS{} expressions $\varphi, \varphi'$, and then comparing to the verifier's answer. 
We utilize \S A2 to construct an evaluation dataset by starting with an \FS{} expression $\varphi$, utilizing a LLM for both \FSNL{} and \NLFS{} to generate the second expression $\varphi'$, where we use a formal verifier to evaluate $\varphi\equiv\varphi'$. We can bootstrap this analysis by using $\varphi$ ($\varphi'$) that are used (generated) in \S A2. 

\subsection{Prompts}
\label{subsec:prompts}
\APPROACH{} includes prompts that allow assessing LLMs in a zero or few-shot capacity. We have engineered our prompts to ensure that (a) all the necessary information for the task is appropriately conveyed, and (b) the LLM is encouraged to not include any context in its output of the \FSNL{} process that could be utilized during \NLFS{}. LLM outputs are automatically checked and parsed using the bundled libraries.  Prompts for each dataset are provided in the supplementary material.

\emph{Prompt Engineering} As stated in Sec.\,\ref{sec:formal_framework}, prompts are extremely important in enabling LLMs to perform tasks well. To ensure that the our results can be viewed in terms of LLM capabilities themselves and not a characteristic of using ``bad'' prompts, we conducted extensive prompt engineering and ensured that at least one LLM can perform \APPROACH{} with $\ge 95\%$ accuracy on a constrained grammar. For example, for logic-based tasks, GPT-4o and Sonnet achieve high accuracy on $3$--SAT using the engineered prompt (results are available in supplementary material). We chose $3$--SAT as the grammar for calibrating logic-based prompts due to its toy structure and since LLMs tend to interpret the formula literal-by-literal. Since the only change between propositional logic, first-order logic and $3$--SAT is the grammar and not the operators themselves, LLM performance for them using the same prompt is a characteristic that is likely isolated to the LLM. We performed similar calibration steps for all datasets. For \S A3, rather than asking for only a yes or no answer, we use Chain-of-Thought (CoT) so that LLMs can utilize their generated outputs to improve their reasoning \citep{wei2022chain}. CoT utilizes more tokens (higher costs), so we also include yes/no prompts.

Additionally, our prompts try to ensure that the LLM does not provide any contextual information between the interpretation and compilation process. This is possible even if the LLM context is cleared between the processes if the LLM encodes relevant contextual information in its output $\Phi$ during \FSNL{} (e.g., copying the formula and then providing a description during interpretation).

\section{Assessment of SOTA LLMs}
\label{sec:assessment}

\begin{figure}
    \centering
    \includegraphics[width=\linewidth]{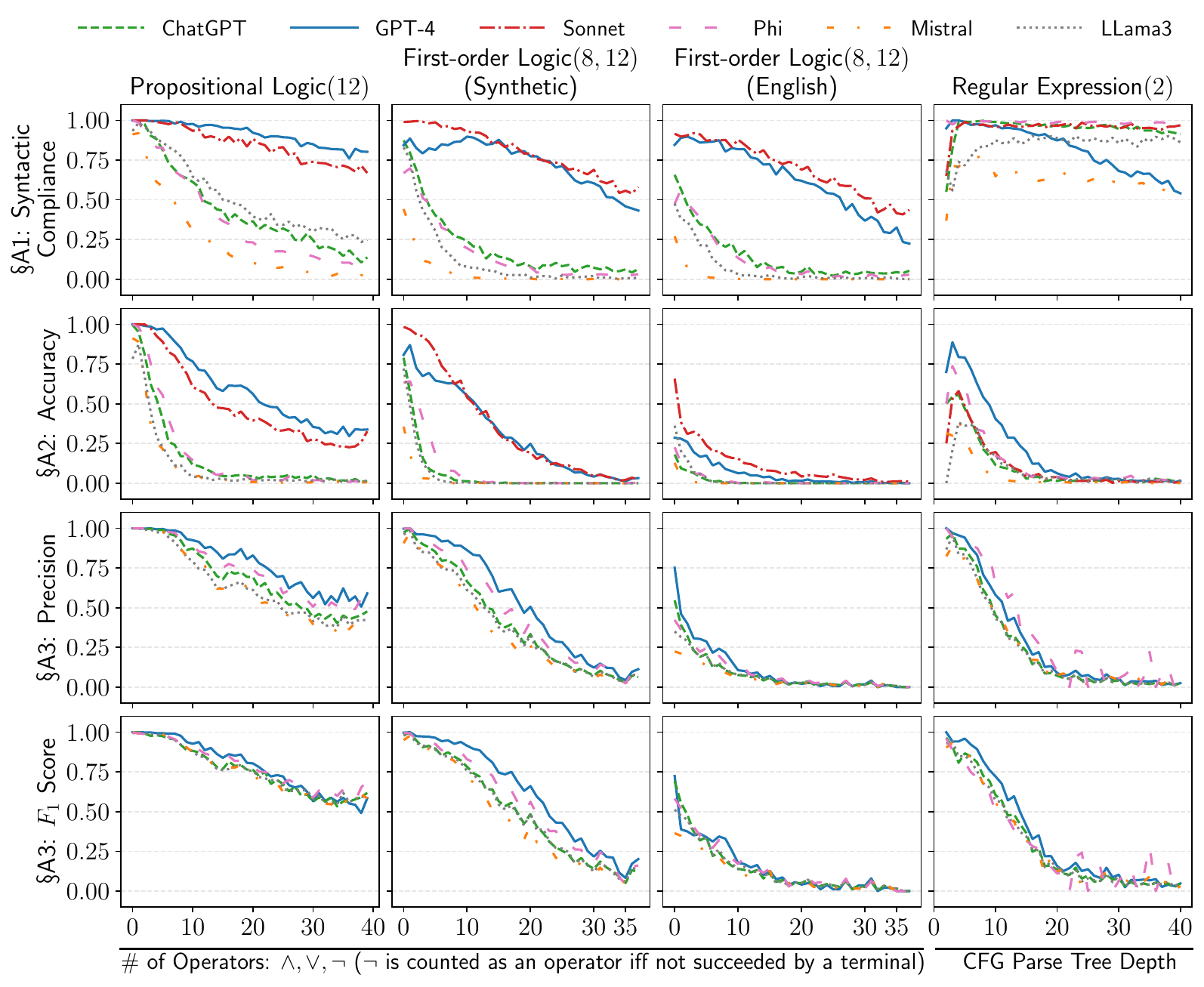}
    \caption{Zero-shot Pass@1 results (avg. over 10 batches, higher values better) from using \APPROACH{} to assess LLMs w.r.t. \S A1, \S A2, \S A3 (Sec.\,\ref{subsec:metrics}) on the packaged datasets. The x-axis represents an increasing order of descriptional complexity. Our prompts, few-shot results, other statistical measures, LLM hyperparameters, compute resources, etc. are included in the supplementary material.}
    \label{fig:results}
    \vspace{-0.23in}
\end{figure}

To motivate research in the area and showcase our framework's strengths, we used \APPROACH{} to evaluate several SOTA closed and open-source LLMs. For paid APIs, we used the latest versions of OpenAI's GPT-4o \citep{gpt-4} and GPT-3.5-turbo \citep{gpt-3.5-turbo} (ChatGPT), and Anthropic's Claude Sonnet\footnote{We did not use Claude Opus since it is 5x more expensive than both GPT-4o and Sonnet} \citep{claude}. We could not utilize Google's Gemini \citep{gemini} due to its extremely constrained rate limiting settings of only 360 requests per minute. Amongst the many open-source LLMs and keeping in mind GPU constraints, we utilized Meta's LLama-3-8B-Instruct \citep{llama3modelcard}, Mistral AI's Mistral-v0.2-7B-Instruct \citep{jiang2023mistral}, and Microsoft's Phi-3-medium-4k-instruct \citep{phi-3} for our evaluation. Our selection of LLMs for evaluating our framework provides a good mix of the current SOTA LLMs in many tasks.

We ran our experiments using both zero-shot and few-shot (2 examples) prompts for all the packaged datasets. This results in $\sim170k$ prompts, thus providing an extremely rich starting point for evaluating other LLMs. Our system can be easily used independently for continual assessment of any LLM. For assessing \S A3, we only ran on OpenAI models to keep costs feasible. Results obtained by using \APPROACH{} are illustrated in Fig.\,\ref{fig:results}. The focus of this paper is to provide an account of the \APPROACH{} framework and its capabilities. Therefore, we provide a short summary of the obtained results and defer providing a detailed description to the supplementary material

\S A1: \emph{Are LLMs syntactically compliant?} As seen in Fig.\,\ref{fig:results}, SOTA LLMs can perform compilation well when the formal syntax has low descriptional complexity (e.g., few operators in propositional logic). But, as the descriptional complexity increases, the performance of LLMs falls drastically.

\S A2: \emph{Can LLMs perform \FTT{} adequately?} Our results show that, except for the prompt calibration task, LLMs cannot perform \FTT{} well as the descriptional complexity increases. Surprisingly, even when using English vocabularies in first-order logic, the performance is worse than using first-order logic with synthetic vocabularies. This is a striking result since the expectation is that LLMs would perform better when presented with first-order logic formulae containing meaningful predicates and objects since LLMs are trained primarily on English datasets. Furthermore, these results are especially concerning as users tend to express formulae in their own vocabulary.

\S A3: \emph{Can LLMs serve as verifiers?} We used the result of formally verifying the ground-truth \FS{} expression $\varphi$ and the LLM output $\varphi'$ generated by $\FS{} \rightarrow \NL{} \rightarrow \FS{}$ to check if replacing the formal verifier with an LLM  yields comparable results. It is clear from Fig.\,\ref{fig:results} that even in this setting, LLMs cannot serve as verifiers for anything but toy expressions (low descriptional complexity), after which precision and $F_1$ scores fall sharply.

\section{Related Work}
\label{sec:related_work}

There is a large body of work for using LLMs w.r.t. formal syntax (e.g., LeanDojo \citep{DBLP:conf/nips/YangSGCSYGPA23} etc). We limit our discussion to approaches that focus on translation of logic and regexes.

\textbf{Datasets for Logic}
RuleTaker~\citep{clark2020transformers} creates a dataset by applying a limited grammar consisting of only conjunctions and disjunctions to generate formulae. This makes it quite limited in the types of formulae it can generate. FOLIO~\citep{han2022folio} has natural language statements with parallel FOL annotations and has expert-written pairs. CLUTTR~\citep{sinha2019clutrr} uses a similar strategy for creating inductive reasoning problems.
RuleTaker, CLUTTR, and FOLIO use human-annotated grammar rules to generate their dataset.

ReClor~\citep{Yu2020ReClor} and
BIG-Bench~\citep{srivastava2023beyond} have hand-coded \NL{} expressions that LLMs use to reason. 
Similarly, ProntoQA~\citep{saparov2023language} is a dataset consisting of logical reasoning queries for FOL. LogicNLI~\citep{DBLP:conf/emnlp/TianLCX0J21} is another FOL dataset that is synthetically constructed through an automated process that generates logical expressions, with placeholders for entities and attributes being systematically substituted to create the final set of problems. LogicBench~\citep{parmar2024logicbench} created a dataset for propositional logic, first-order logic, and non-monotonic reasoning. 
ReClor, BIG-Bench, LogicNLI, LogicBench, and ProntoQA use logic rules to create datasets but do not use verifiers to ensure that the dataset uses accurate natural language corresponding to the logic rules and hence are prone to inconsistencies.

LogicLLaMa \citep{DBLP:journals/corr/abs-2305-15541} is an LLM trained specifically for the \emph{NL$\rightarrow$FS} task. Inorder to evaluate its efficacy, the authors proposed MALLS, a custom generated dataset that can be used to evaluate the efficacy of LogicLLaMa. Their approach does not provide an automatic way to test the efficacy of LLMs in the translation task and requires hand-coded datasets. Moreover, their evaluation metric considers logical equivalence using truth tables which can be inefficient. Autoformalization \citep{DBLP:conf/nips/WuJLRSJS22} uses pre-trained LLMs together with few-shot prompting to convert NL descriptions to formal syntax. This approach requires expert prompting and relevant examples to boost performance. Furthermore, this approach relies on existing, annotated datasets and thus can be susceptible to LLMs having memorized the answers. \citet{DBLP:journals/corr/abs-2302-05128} use few-shot prompting to translate NL descriptions to PDDL goals and assess them using manually designed, domain-dependent metrics. Moreover, their approach cannot perform \emph{FS$\rightarrow$NL} translations.

These approaches either (a) need human annotators, (b) do not test both directions of \FTT{}, or (c) use static datasets. In contrast, \APPROACH{} is handsfree, tests both directions, and is scalable.

\textbf{LLMs that interpret NL}
Most common paradigms for NL to first-order logic translation use rule-based approaches~\citep{abzianidze2017langpro,zettlemoyer2005Learning,bos2005recognising} but these approaches are plagued with issues of scalability and performance on real-world data. A few approaches try to alleviate these issues by using neural architectures~\citep{cao2019semantic,lu2022parsing}.
It has also been shown that LLMs when trained on programming code demonstrate improved reasoning as they frame reasoning as generation of code~\citep{chen2021evaluating}.
CloverBench~\citep{sun2023clover} dataset is intended to check equivalence for code standard in a pair of input-output codes but require human input.
These approaches either depend on an expert to verify the generated FOL or have to keep an expert in the loop while verifying the output and thus do not scale. \APPROACH{} uses a verifier to check the output w.r.t. the ground-truth efficiently and without requiring any human intervention.

\section{Closing Remarks}
\label{sec:conclusions}

We now provide our conclusions followed by a discussion of some limitations and future work following which we remark on some ethical considerations of our work.

\textbf{Code Access and Extensibility} Our code is hosted at \url{https://github.com/AAIR-lab/auto-llm-assessment} and is open-source under the GNU GPLv3 license \citep{gnu_license}. Our code includes documentation on how to run our experiments and also adding new LLMs and datasets. We aim to continually update our code by adding new features, grammars, LLM support etc.

\textbf{Conclusions} This paper introduced \APPROACH, a new benchmark and dynamic datasets for autonomous assessment of LLM capabilities in the formal translation task. Our approach for dataset synthesis avoids the pitfalls that befall current approaches in that they can be scaled arbitrarily without requiring any human-labelling. Our approach for assessment evaluates \FTT{} in both directions without any human-intervention and provides accurate results by using verifiers to guarantee correctness over all inputs. Our framework is easily extensible and provides several prepackaged datasets to quickly assess LLMs. Furthermore, our evaluation indicates that SOTA LLMs are not performant in this task.

\textbf{Broader Impact} This work introduces a way to automatically assess both directions of formal translation while automatically scaling datasets. Applications of this work could be used by researchers or industries to robustly evaluate the performance of LLMs in this task. These results could be used to determine suitability and safety of using LLMs in formal translation.

\textbf{Limitations and Future Work}  Our framework assumes that intermediate \NL{} output $\Phi$ must be correct if $\varphi \equiv \varphi'$ for a given input $\FS{}$ $\varphi$. It is possible that a hallucinated $\NL$ description $\Phi$ might be compiled by the LLM to $\varphi' = \kappa_{\LLM{}}(\Phi)$ s.t. $\varphi \equiv \varphi'$ even though $\kappa(\Phi) \not \equiv \varphi$. One way to mitigate this is to use the top-$k$ responses from the LLM in the \NLFS{} step and compare each of the formulae. A high accuracy across the $k$ responses would lead to increased confidence of $\NL$ being correct, albeit at additional cost for paid APIs. One interesting extension of current work is to utilize the $\lambda$-calculus to further expand the domains that could be represented using CFGs. Our current framework uses \FS{} as inputs. It is possible to use our framework as-is to generate \NL{} $\Phi, \Phi'$ as input and output by providing appropriate verifiers. However, it is quite difficult to find good verifiers that check for semantic similarity, i.e., $\Phi \equiv \Phi'$. 
Finally, it is well-known that first-order logic is not decidable \citep{DBLP:journals/x/Turing37}. Thus, \FS{} verifiers must be loaded with the appropriate timeout or grammars used for synthesizing datasets must only generate strings from decidable fragments of the formal syntax.

\textbf{Threats to Validity} Our reported results for paid APIs are dependent on the model checkpoints used to be available. Most paid APIs provide a disclaimer that model weights can be updated for the same version number without notice. As is the case with several evaluation methodologies for LLMs, one must be careful when interpreting results and use pass@k and other detailed measures such as standard deviations to increase confidence. We report pass@1 due to the high-cost of pass@k for paid APIs and cluster time for open-source LLMs but do report standard deviations across 10 runs in the supplementary material. \S A3 utilized the input \FS{} $\varphi$ and the LLM response $\varphi'$ in \S A2 and the verifier answer, $\varphi \equiv \varphi'$, as input, leading to an imbalanced dataset of negative examples at high descriptional complexity due to low precision. However, positive \FS{} examples $\varphi''$ can be easily created from $\varphi$ by using the verifier to \textit{simplify} them to generate $\varphi''$ s.t. $\varphi \not= \varphi''$ and $\varphi \equiv \varphi''$. As is the case with many existing works, our approach assumes the soundness of the verifier programs and parsing libraries used. Software bugs in the verifier program or parsing libraries could cause false signals to be output by \APPROACH{}.

\textbf{Ethical Considerations} Our work involves the use of LLMs for formal translation. Naturally, it is imperative to ensure that appropriate guardrails are in place to prevent offensive content from being generated and/or displayed. We do not use any personally identifiable information in \APPROACH{}.

\section*{Acknowledgements}
This work was supported in part by the ONR grant N00014-21-1-2045, NSF grant IIS 1942856, the Open AI Researcher Access Grant, and Arizona State University's GPSA Jumpstart Research Grant.

{
\bibliographystyle{plainnat}
\bibliography{neurips_data_2024}
}

\newpage

\appendix

\section{Appendix Organization}

Access to the code is provided at \url{https://github.com/AAIR-lab/auto-llm-assessment} under the GNU GPLv3 license. The datasets are in the code repository and at \url{https://doi.org/10.5281/zenodo.11628901} under the Creative Commons license CC BY 4.0. Instructions to reproduce our results are available in the code's README file.

The appendix is organized as follows. Appendix \ref{appendix:dataset_info} discusses the datasets released with this benchmark. Appendix \ref{appendix:dataset_gen} provides the algorithm used for dataset generation. Appendix
\ref{appendix:prompt_tuning} discusses prompt tuning and validating our prompts on 3SAT. Appendix \ref{appendix:dataset_parameters} provides the parameters we used when generating the five datasets discussed in the paper. Appendix \ref{appendix:experimental_setup} provides additional information on our experimental setup, including the computational resources used. Appendix \ref{appendix:prompting} discusses the prompts and provides examples. Appendix \ref{appendix:main_paper_results} is our detailed analysis of the empirical results from the main paper. Appendix \ref{appendix:std_dev_eval} discusses an experiment we ran to evaluate the standard deviation error. Appendix \ref{appendix:add_zero} includes additional results from our zero-shot prompting experiments using other metrics for categorization. Finally, Appendix \ref{appendix:few_shot} includes the results of our experiment evaluating few-shot prompting compared to zero-shot.

\section{Dataset Information}
\label{appendix:dataset_info}

In this section, we discuss the datasets we have released with this benchmark. We provide the datasets under the Creative Commons CC BY 4.0 license. 

\subsection{Dataset Storage and Maintenance Plans}

The datasets are in the code repository and are also permanently hosted at \url{https://doi.org/10.5281/zenodo.11628901}. The original dataset used to reproduce the results in the paper will not be changed. The other dataset (discussed in the next section) is provided for future research. If we are required to update this dataset, the previous versions will be maintained in the GitHub repository. 

\subsection{Dataset Usage Guidelines}

We are releasing two datasets with this paper: the five datasets discussed in the paper used for our empirical evaluation and the resulting dataset constructed during the evaluation.

\textbf{Generated Dataset: } For evaluating LLM's ability to interpret formal syntax and compile formal syntax from natural language, we produced 5 datasets discussed further in Section\,\ref{sec:datasets}. These datasets can be used with \APPROACH{} to draw a fair comparison with the empirical evaluation performed in this paper and can be used to reproduce our results. Instructions in our code base make it straightforward to evaluate different LLMs on this problem using the datasets.

\textbf{\APPROACH{} Dataset: } As a byproduct of our empirical evaluation using the datasets generated, we have created a dataset containing the intermittent steps of the \APPROACH{} process: \FSNL{}, \NLFS{}, equivalence verification, and formal verification. For \FSNL{} and \NLFS{}, this includes the full conversations produced with each LLM evaluated. We are releasing this dataset to allow for further evaluation and future research. For example, the \FSNL{} results can be used to calculate the generated explanations' BLEU scores. Additionally, a new LLM could be trained on the positive and negative labels to improve its ability at this type of task. Furthermore, since a new dataset can be generated, \APPROACH{} can be used to evaluate this new LLM without the risk of overfitting.

\section{Dataset Generation}
\label{appendix:dataset_gen}

In this section, we provide the algorithm for generating formal syntax (FS) expressions and show that it can generate all possible expressions from the grammar and vocabulary.

Our approach, \APPROACH{}, generates datasets by constructing a context-free grammar (CFG) tree using the grammars shown in Fig.\,\ref{fig:fig2}. Since it is intractable to generate the full tree, we control the branching factor and randomly expand the branches of this tree to generate formulae.

\begin{algorithm}[H]
\caption{Dataset Generation}
\label{alg:ips}
\begin{algorithmic}[1]
\State \textbf{Inputs:} CFG $\mathcal{G}$, vocabulary $\mathcal{V}$, branching factor $n$, tree depth $depth$, sample count $sample\_count$, and categorization metric $m$.
\State \textbf{Outputs:} set of FS expressions $\overline{\varphi}$
\State $\mathcal{N} \leftarrow \{0:[None]\}, \mathcal{N}_t \leftarrow \langle\rangle$
\For {$d=1,2,\ldots,depth$}
\State $\mathcal{N}'\leftarrow SampleN(\mathcal{N}[d-1],n)$
\For {$\nu \in \mathcal{N}'$}
\State $\mathcal{N}_\nu,\mathcal{T}_\nu 
\leftarrow GenerateNChildren(\nu,\mathcal{G},n)$
\State $\mathcal{N}[d] \mathrel{+}= \mathcal{N}_\nu$
\State $\mathcal{N}_t \leftarrow \mathcal{N}_t \cup \mathcal{T}_\nu$
\EndFor
\EndFor
\State $M \leftarrow CategorizeExpressionsIntoDict(\mathcal{N}_t,m)$
\State $\overline{\varphi} \leftarrow \langle\rangle$
\For {$k \in keys(M)$}
\State $M_k \leftarrow SampleCFGExpressions(M[k],sample\_count)$
\State $\overline{\varphi}_k \leftarrow BuildFSExpressions(M_k,\mathcal{V})$
\State $\overline{\varphi} \leftarrow \overline{\varphi} \cup \overline{\varphi}_k$
\EndFor
\State \textbf{Return: } $\overline{\varphi}$
\end{algorithmic}
\label{alg: alg1}
\end{algorithm}
\vspace{-0.1in}

The dataset generation algorithm is shown in Algorithm\,\ref{alg: alg1}. This algorithm constructs a CFG tree by maintaining non-terminal nodes at each tree level (
$\mathcal{N}$) and all the leaf nodes ($\mathcal{N}_t$), where each terminal node represents a completed CFG expression (line 3). For generating nodes at a certain level in the tree, $n$ nodes from the previous level are sampled (line 5). Each node is branched $n$ times using the CFG to produce nodes at the current tree level, and all the leaf nodes are collected (lines 7 through 9). As a result, by iteratively performing this process for each tree level, we obtain a set of leaf nodes (CFG expressions).

The leaf nodes are then categorized based on the specified metric (e.g., tree depth, number of operators, etc.) (line 12). For each metric value, a fixed number of CFG expressions corresponding to that value are sampled (line 15). Using the vocabulary, an FS expression is constructed from each CFG expression (line 16). Consequently, the final dataset of FS expressions contains an equal number for each metric value (line 17). This set of FS expressions is the final result produced by the algorithm (line 19).

The vocabulary is fixed in length, with a hyperparameter controlling the number of unique propositions for propositional logic. Similarly, for first-order logic, the number of unique variables, constants, and predicates are also hyperparameters. Also, regular expressions have a hyperparameter controlling the alphabet size. When these expression components are needed for building the FS expression, the exact one is selected using uniform random selection. In the special case of first-order logic predicates, the grounded predicate is generated by randomly selecting a predicate and then selecting constants depending on the predicate's arity. In the case of the arbitrary vocabulary, the arity for a predicate is randomly assigned. To add variables, each constant has a certain probability of being replaced by a variable.

\textbf{Guaranteed Expression Coverage} 
The dataset generator (Algorithm\,\ref{alg: alg1}) is guaranteed to generate all possible formal syntax expressions that can be produced with the given grammar and vocabulary. Let $\varphi$ be an FS expression that can be constructed using the rules from CFG  $\mathcal{G}$ and the vocabulary $\mathcal{V}$. Note that $\varphi$ corresponds to a CFG expression $\varphi_{CFG}$, derived by substituting the vocabulary with the CFG symbols. Due to uniform selection, the probability of $\varphi$ being generated from $\varphi_{CFG}$ is greater than zero. Furthermore, the CFG expression represents a leaf node in the CFG tree that can be reached by applying the CFG rules in a specific sequence. Due to the random sampling of rules at each node, there is a non-zero probability of generating this particular path in the tree. Thus, $\varphi$ can be generated using the dataset generator algorithm.

\section{3-SAT Prompt Tuning}
\label{appendix:prompt_tuning}

In this section, we discuss the KSAT results used to fine-tune the prompts.

\begin{figure}[h]
    \centering
    \includegraphics[width=\linewidth]{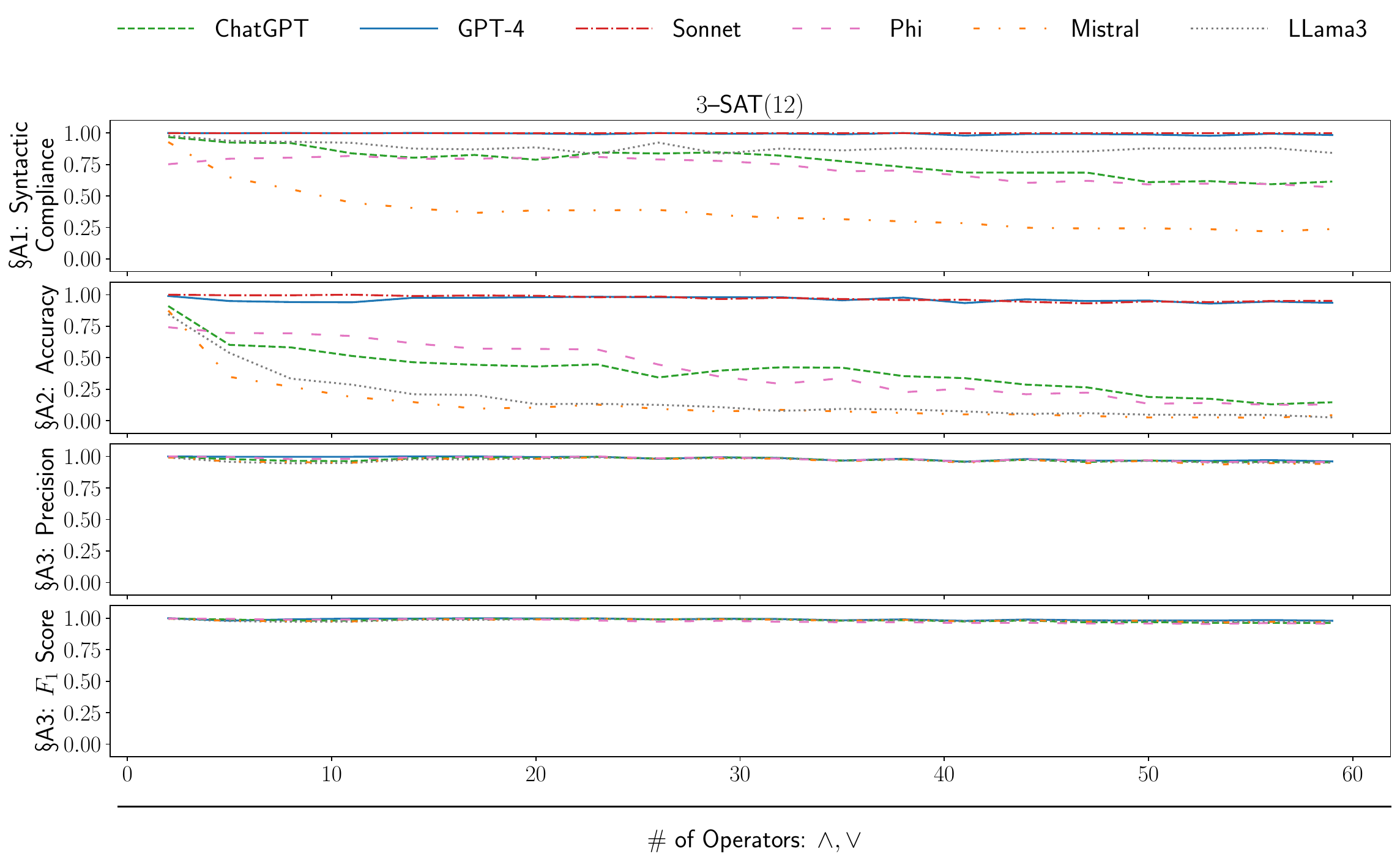}
    \caption{Zero-shot Pass@1 results (avg. over 10 batches, higher values better) for 3-SAT from using \APPROACH{} to assess LLMs w.r.t. \S A1, \S A2, \S A3 (Sec.\,\ref{subsec:metrics}) on the packaged datasets. The x-axis is the \# of operators.}
    \label{fig:ksat_results}
\end{figure}

We tested several prompts for 3-SAT to verify that our prompts are sufficient to prompt the LLM to produce high-quality explanations for \FSNL{} and accurate FS expression for \NLFS{}. Additionally, we verified that the equivalence verification prompt prompted the LLMs to give an accurate yes-or-no answer. The performance of all six LLMs on 3-SAT for \S A1, \S A2, and \S A3 are shown in Figure\,\ref{fig:ksat_results}.

The best-performing models we tested (GPT-4o and Claude Sonnet) achieved nearly perfect syntactic compliance, accuracy, and equivalence verification even as the number of operators increased. This proves that the prompts we used in our experiments are sufficient for prompting the model for solving \S A1, \S A2, and \S A3.

For the other LLMs tested, syntactic compliance and accuracy diminished as the number of operators increased. However, when evaluating the equivalence of GPT-4o results, all LLMs achieved near-perfect accuracy regardless of operator number. Due to most of GPT-4o results being positive cases, the results imply that LLMs can evaluate two equivalent 3-SAT expressions correctly.

\section{Dataset Generation Hyperparameters}
\label{appendix:dataset_parameters}

In Table\,\ref{tab:dataset_params}, we provide the hyperparameters used to generate the five datasets.

\begin{table}[h!]
    \centering
    \begin{tabular}{l l @{}c p{0.38\columnwidth}}
         \toprule
         \textbf{Parameter Type} & \textbf{Hyperparameter} & \textbf{Value} & \textbf{Description} \\ \midrule
         \multirow{3}{*}{\textbf{General}} &
         \texttt{depth} \CC & 40 \CC & Maximum depth of the CFG tree. \CC \\
         & \texttt{n} & 200 & Branching factor of produced CFG tree. \\
         & \texttt{sample\_count} \CC & 50 \CC & Number of CFGS for each metric value to select. \CC \\ \midrule
         \multirow{6}{*}{\textbf{First-Order Logic}} &
          \texttt{free\_variable\_prob} & $0.25$ & Probability of a constant being replaced by a variable. \\
          & \texttt{max\_free\_variables} \CC & $\infty$ \CC & Maximum number of unique variables.\CC \\
         & \texttt{max\_predicate\_arity} & 2 & Maximum predicate arity. \\ 
         & \texttt{min\_predicate\_arity} \CC & 1 \CC & Minimum predicate arity. \CC \\ 
         & \texttt{num\_objects} & 12 & Number of unique constants. \\
         & \texttt{num\_predicates} \CC & 8 \CC & Number of unique predicates. \CC \\ \midrule
         \textbf{Propositional Logic} &
         \texttt{num\_propositions} & 12 & Number of unique propositions. \\\midrule
         \textbf{Regular Expression}  & 
          \texttt{alphabet\_size} \CC & 2 \CC & Alphabet size. \CC \\ \bottomrule\\
    \end{tabular}
    \caption{Hyperparameters used for producing the five datasets.}
    \label{tab:dataset_params}
\end{table}

\section{Experimental Setup}
\label{appendix:experimental_setup}

In this section, we will provide the details of our experimental setup for generating the datasets and running \APPROACH{} for evaluating each LLM's performance.

We ran our experiments using Python 3.10.13 with package versions shown in Table\,\ref{tab:python_version}. We also repackaged Prover9 \citep{prover9-mace4} to improve performance where this repackaged version can be found in our code base.   

\begin{wraptable}{r}{0.33\textwidth}
    \vspace{-0.4cm}  
\begin{minipage}[r]{0.33\textwidth}
    \centering
    \rowcolors{2}{gray!13}{}
    \begin{tabular}{l l }
         \toprule \textbf{Python Package} & \textbf{Version} \\ \midrule
         \texttt{openai} & 1.30.3 \\
         \texttt{nltk} & 3.8.1 \\
         \texttt{tqdm} & 4.66.4 \\
         \texttt{anthropic} & 0.26.1 \\
         \texttt{backoff} & 2.2.1 \\
         \texttt{tiktoken} & 0.6.0 \\
         \texttt{transformers} & 4.41.1 \\
         \texttt{Faker} & 25.2.0\\
         \texttt{networkx} & 3.3 \\ \bottomrule
    \end{tabular}
    \caption{Python package versions used for empirical evaluation.}
    \label{tab:python_version}
\end{minipage}
\end{wraptable}

\textbf{Dataset Generation: } We generated five datasets using the dataset generation algorithm with the hyperparameters shown in Table\,\ref{tab:dataset_params} using the number of operators as the categorization metric for all but regular expression, where we used CFG tree depth. We generated 10 batches for each dataset, resulting in approximately 20k samples for each dataset with an equal distribution for each operator number.

\textbf{Evaluating and Verification: } The closed-source models (GPT-3.5-turbo, GPT-4o, and Claude Sonnet) were accessed using their API using a temperature of 0.1. Claude Sonnet had the max tokens set to 1024. The open-source models LLama-3-8B-Instruct and Mistral-v0.2-7B-Instruct were locally hosted on a server with a 13th Gen Intel(R) Core(TM) i9-13900K and Nvidia RTX 4090 GPU using the model's default parameters with a temperature of 1. Similarly, Phi-3-medium-4k-instruct was locally hosted on a server using a Nvidia A100-XM4-80GB GPU. Verification was performed on an AMD EPYC machine with 128 cores.

\section{Prompting}
\label{appendix:prompting}

In this section, we provide the prompts used for the zero-shot and few-shot prompting experiments.

The prompt for each dataset type provides the LLM with information on the problem type and the vocabulary. For \FSNL{}, we prompt the model to produce just a natural language description. We also provide the list of objects, predicates, propositions, and free variables in the formal syntax expression. For \NLFS{}, the LLM is prompted to provide just the formal syntax expression using the natural language description. Two examples are provided in the \FSNL{} and \NLFS{} prompts for few-shot prompting.

For \S A3, the prompt used for using an LLM to verify the equivalence of two formulae tells the LLM about the type of datasets (e.g., propositional logic, first-order logic, and regular expression). Using chain-of-thought prompting, the model is prompted to provide an explanation before giving a yes-or-no answer in a format we can parse.  Below are examples of the exact prompts used.

\begin{mdframed}[style=PVFrame,frametitle={Few Shot FOL Prompt NL to Formal}]
 [TASK] \newline  Your task is to convert a first-order logic formula, appearing after [FORMULA], to a natural description that represents the formula. Only natural language terms are allowed to be used and do not copy the formula in your description. Your description should allow one to reconstruct the formula without having access to it, so make sure to use the correct names in your description. Explicitly describe the predicates. You may use terms verbatim as specified in the vocabulary below. \hfill \newline \newline [EXAMPLE 1]\newline $(\neg p2 \lor p1 \lor \neg p2)$ \newline Disjunctive predicate logic expression consisting of three components: the negation of a proposition labeled p2, the proposition p1, and again the negation of p2.\hfill \newline \newline [EXAMPLE 2]\newline $(\neg \neg p2 \land  \neg (p3 \lor p1))$\newline The expression asserts that p2 is not false while both p3 and p1 are not true.\hfill \newline \newline [VOCABULARY] \newline $\lor$ represents disjunction \newline $\land$  represents conjunction \newline $\neg$  represents negation \newline ( and ) represent parentheses \newline propositions can be used verbatim\newline predicates can be used verbatim \newline $\forall  <x1> <x2> ... <xn>.$ represents universal quantification with x1... representing free variables \newline $\exists <x1> <x2> ... <xn>.$ represents existential quantification with x1... representing free variables\newline The objects are: $p5, x1$ \newline The parameterized predicates are: $pred3(?p0,?p1)$ \newline The free variables are: $x1$ \newline \newline [FORMULA] \newline $\forall x1$ $pred3(p5, x1)$
\end{mdframed}

\begin{mdframed}[style=PVFrame,frametitle={Few Shot FOL Prompt Formal to NL}]
[VOCABULARY] \newline Use $\lor$ to represent disjunction \newline Use $\land$  to represent conjunction \newline Use $\neg$  to represent negation \newline Use ( and ) to represent parentheses \newline Use $\forall$  <free\_variable\_list> to represent universal quantification \newline Use $\exists$ <free\_variable\_list> to represent existential quantification \newline The <free\_variable\_list> consists of a sequence of space separate free variables with the last variable immediately followed by a period. Examples: (1) all x1 x2. (2) exists x4. \newline Use <predicate>(<parameter\_list>) to represent predicates (Names and parameters are provided in the description) \newline  \newline [TASK] \newline Your task is to interpret the natural language (NL) description of a first-order logic formula and represent it as formal syntax using the vocabulary specified in the [VOCABULARY] block above. Only output the formula and no other text. The NL description appears immediately following the [NL DESCRIPTION] tag. \newline  \newline [EXAMPLE 1] \newline Disjunctive predicate logic expression consisting of three components: the negation of a proposition labeled p2, the proposition p1, and again the negation of p2. \newline $(\neg p2 \lor p1 \lor \neg p2)$ \newline \newline [EXAMPLE 2] \newline The expression asserts that p2 is not false while both p3 and p1 are not true. \newline $(\neg \neg p2 \land  \neg (p3 \lor p1))$ \newline \newline [NL DESCRIPTION] \newline For all objects labeled x1, the predicate pred3 holds true with parameters p5 and x1.
\end{mdframed}

\begin{mdframed}[style=PVFrame,frametitle={Few Shot Regex Formal to NL Prompt}]
  [TASK] \newline Your task is to convert the regular expression appear after [REGEX], to a natural description that represents the regular expression. Only natural language terms are allowed to be used and do not copy the regular expression in your description. Your description should allow one to reconstruct the regular expression without having access to it, so make sure to use the correctly account for scoping. You may use terms verbatim as specified in the vocabulary below. \newline  \newline [VOCABULARY] \newline you may use symbols from the vocabulary \newline you can use * \newline  \newline [EXAMPLE 1] \newline (1*)0* \newline The regex matches strings that starts with any number (including none) of the digit '1', followed by any number (including none) of the digit '0'. \newline \newline [EXAMPLE 2] \ \newline (01*) \newline The regex matches strings that begin with a '0' followed directly by any number (including none) of '1's. \newline \newline [FORMULA]  \newline 0
\end{mdframed}

\begin{mdframed}[style=PVFrame,frametitle={Few Shot Regex NL to NL Formal}]
[VOCABULARY] \newline Use * to represent zero or more duplications of the same expression \newline Use ( and ) to represent parentheses \newline  \newline [TASK] \newline Your task is to interpret the natural language (NL) description of a regular expression and represent it as formal syntax using the vocabulary specified in the [VOCABULARY] block above. Only output the regular expression and no other text. The NL description appears immediately following the [NL DESCRIPTION] tag. \newline  \newline [EXAMPLE 1] \newline The regex matches strings that starts with any number (including none) of the digit '1', followed by any number (including none) of the digit '0'. \newline (1*)0* \newline  \newline [EXAMPLE 2] \newline The regex matches strings that begin with a '0' followed directly by any number (including none) of '1's. \newline (01*) \newline  \newline [NL DESCRIPTION] \newline The regex matches strings that start with the digit '0'.
\end{mdframed}

\begin{mdframed}[style=PVFrame,frametitle={Propositional Zero Shot Formal to NL}]
  [TASK] \newline Your task is to convert a propositional logic formula, appearing after [FORMULA], to a natural description that represents the formula. Only natural language terms are allowed to be used and do not copy the formula in your description. Your description should allow one to reconstruct the formula without having access to it, so make sure to use the correct names in your description. Explicitly describe the predicates. You may use terms verbatim as specified in the vocabulary below. \newline  \newline [VOCABULARY] \newline $\lor$ represents disjunction \newline $\land$  represents conjunction \newline $\neg$  represents negation \newline ( and ) represent parentheses \newline propositions can be used verbatim \newline The propositions are: p5, p12, p4 \newline  \newline [FORMULA] \newline $(p5 \lor \neg p12 \lor \neg p4)$
\end{mdframed}

\begin{mdframed}[style=PVFrame,frametitle={Propositional Zero Shot NL to Formal}]
  [VOCABULARY] \newline Use $\lor$ to represent disjunction \newline Use $\land$  to represent conjunction \newline Use $\neg$  to represent negation \newline Use ( and ) to represent parentheses \newline  \newline [TASK] \newline Your task is to interpret the natural language (NL) description of a propositional logic formula and represent it as formal syntax using the vocabulary specified in the [VOCABULARY] block above. Only output the formula and no other text. The NL description appears immediately following the [NL DESCRIPTION] tag. \newline \newline [NL DESCRIPTION] \newline A disjunctive statement involving three propositions: p5, the negation of p12, and the negation of p4.
\end{mdframed}

\begin{mdframed}[style=PVFrame,frametitle={Verification Prompt}]
Your task is to say whether two  First Order Logic are equivalent. The first formula will appear right after [FORMULA 1] and the second after [FORMULA 2]. \newline
        Give an explanation followed by a yes or no answer. The answer must show up at the end with the format "[Answer]" followed by either a yes or no. \newline \newline
        [Formula 1] \newline 
        $\exists x1.\neg pred5(p7)$ \newline \newline 
        [Formula 2] \newline
        $\exists p7.\neg pred5(p7)$
\end{mdframed}

\section{Analysis of Main Paper Results}
\label{appendix:main_paper_results}

In this section, we analyze the main empirical results of the paper. Our results clearly show that current SOTA LLMs are not performant in the $NL\leftrightarrow FS$ task. As the expression complexity increases, the syntactic compliance, accuracy, and ability to verify equivalence diminishes. We describe some of the errors that cause the low accuracy for propositional logic, first-order logic, and regular expressions.

\subsection{Propositional Logic Results}

\begin{table}[h]
    \centering
    \begin{tabular}{l l l}
         \toprule \textbf{FS} & \textbf{\FSNL{}} & \textbf{\NLFS{}} \\ \midrule
        $(\neg p11 \land \neg p8)$ \CC & \begin{tabular}{@{} l@{}}
        The natural description of the formula is: \\"It is not the case that both proposition p11 \\ and proposition p8 are true."
        \end{tabular} \CC & $(\neg (p11 \land p8))$ \CC \\ 
                $(\neg p10 \land (\neg p9 \lor p7))$ & \begin{tabular}{@{} l@{}}
        The natural description of the formula is: \\"It is not the case that proposition p10 is true \\and either proposition p9 is false or \\proposition p7 is true."
        \end{tabular} & $\neg (p10 \land (\neg p9 \lor p7))$ \\ 
        $(\neg p3 \land \neg p7)$ \CC & \begin{tabular}{@{} l@{}}
        Here is the natural description of the given \\formula: "The statement says that p3 is not \\true and p7 is also not true."
        \end{tabular} \CC & $(\neg p3 \lor \neg p7)$ \CC \\
        $(\neg\neg p2 \lor p3)$ & \begin{tabular}{@{} l@{}}
        The formula states that either the proposition \\p2 is true or the proposition p3 is true. \\Additionally, it specifies that the proposition \\p2 is not false.
        \end{tabular} & $(p2 \lor p3) \land \neg\neg p2$ \\
        $(\neg p2 \land p5 \land \neg p6)$ \CC & \begin{tabular}{@{} l@{}}
        The formula describes a situation where the \\ first proposition is false, the second \\proposition is true, and the third proposition \\is false.
        \end{tabular} \CC & $(\neg p \land q \land \neg r)$ \CC \\
        \bottomrule
    \end{tabular}
    \caption{Examples of errors the evaluated LLMs made while evaluating with \APPROACH{} for the propositional logic dataset.}
    \label{tab:plog_error_examples}
\end{table}

\textbf{\FSNL{} Errors:} A common error was the LLM failed to describe the proposition names. Another was the LLM failing to provide a complete description of the formula. For example, GPT-3.5-turbo often described portions of the expression based on what propositions and operators it contained. A common issue with GPT-4o, one of the best models, is that it often uses different propositional symbols (see example 5 in Table\,\ref{tab:plog_error_examples}). Finally, we also observed hallucinations that were more common when the LLM attempted to simplify the original formula (see example 4 in Table\,\ref{tab:plog_error_examples}). These interpretation errors resulted in the original meaning of the expression being lost.

\textbf{\NLFS{} Errors:} A common problem was the LLM ignored the prompt to give just the formal syntax expression. For example, the LLM would describe the formula after providing it. Additionally, there were often syntactic issues where the description was not fully translated into a formula or the parentheses did not match. An interesting result is that the LLMs struggled to place the negation operator in the correct location. For example, GPT-4o often describes $\neg p \land\neg p$ as predicate $p$ "negated twice and combined" but failed to regenerate the original formula properly with this description.

\subsection{First-Order Logic Results}

\begin{table}[h]
    \centering
    \begin{tabular}{l l l}
         \toprule \textbf{FS} & \textbf{\FSNL{}} & \textbf{\NLFS{}} \\ \midrule
    \begin{tabular}{@{} l@{}}
        $ (\neg pred8(p10) \land pred8(p5)$ \\ $ \land pred6(p8))$
        \end{tabular} \CC & \begin{tabular}{@{} l@{}}
        The formula states that it is not the case\\ that the predicate pred8 holds for the \\object p10, and the predicate pred8 \\holds for the object p5, and the \\predicate pred6 holds for the object p8, \\all conjoined together.
        \end{tabular} \CC &     \begin{tabular}{@{} l@{}}
        $\neg (pred8(p10)$ \\$\land pred8(p5)$ \\ $ \land pred6(p8))$
        \end{tabular}  \CC \\
    \begin{tabular}{@{} l@{}}
        $ \exists x1.\neg pred2(p4)$
        \end{tabular}  & \begin{tabular}{@{} l@{}}
        Here is the natural description of the \\given formula: "There exists an x1 such \\that p4 is not an instance of pred2.
        \end{tabular}  &     \begin{tabular}{@{} l@{}}
        $\exists x1. \neg pred2(x1)$
        \end{tabular} \\
    \begin{tabular}{@{} l@{}}
        $ \forall x1.\neg\neg pred3(p5)$
        \end{tabular} \CC & \begin{tabular}{@{} l@{}}
        For all x1, it is not the case that \\pred3(p5) is false.
        \end{tabular} \CC &     \begin{tabular}{@{} l@{}}
        $\forall x1. \neg(pred3(p5)$\\ $ \lor \neg pred3(p5))$
        \end{tabular}  \CC \\

    \begin{tabular}{@{} l@{}}
        $ pred2(p3,p5)$
        \end{tabular}  & \begin{tabular}{@{} l@{}}
        There exists a relationship between the  \\objects p3 and p5 such that the predicate \\pred2 holds true for these objects.
        \end{tabular}  &     \begin{tabular}{@{} l@{}}
        $\exists p3$ $p5. pred2(p3, p5)$
        \end{tabular} \\

        \bottomrule
    \end{tabular}
    \caption{Examples of errors the evaluated LLMs made while evaluating with \APPROACH{} for the first-order logic dataset.}
    \label{tab:fol_error_examples}
\end{table}

\textbf{\FSNL{} Errors:} The same lack of detail for formula explanations was observed, resulting in incorrect formulas being generated. A significant source of errors was the LLM rewording its explanation causing confusion when regenerating. For first-order logic with English predicate and object names, all the LLMs often failed to describe the predicates and objects, resulting in the LLMs failing to regenerate the same predicates and objects. For example, it was quite common for the predicate and object names to be swapped (e.g., $stew(Deane)$ would become $Deane(stew)$).

\textbf{\NLFS{} Errors:} Beyond the errors observed in propositional logic, the most common mistake made during compilation was the LLM confusing constants with variables (see example 2 in Table\,\ref{tab:fol_error_examples}). Additionally, the LLMs often messed up the predicate arity. Mistral often used $=$ and $\neq$ operators with the quantifiers, which was not needed for any of the expressions. Similarly, the LLMs would often use their own grammar instead of the one provided in the prompt.

\subsection{Regular Expression Results}

\begin{table}[h]
    \centering
    \begin{tabular}{l l l}
         \toprule \textbf{FS} & \textbf{\FSNL{}} & \textbf{\NLFS{}} \\ \midrule
    \begin{tabular}{@{} l@{}}
        $1^*0$
        \end{tabular} \CC & \begin{tabular}{@{} l@{}}
        The pattern consists of one or more occurrences of the digit \\"1" followed by exactly one occurrence of the digit "0".
        \end{tabular} \CC &     \begin{tabular}{@{} l@{}}
        $(1^*)10$
        \end{tabular}  \CC \\
    \begin{tabular}{@{} l@{}}
        $(1^*)^*0$
        \end{tabular} & \begin{tabular}{@{} l@{}}
        The regular expression describes a sequence that consists of \\ zero or more repetitions of the digit '1', followed by exactly \\ one '0'. This entire pattern can repeat any number of times, \\ including zero times.
        \end{tabular} &     \begin{tabular}{@{} l@{}}
        $((1^*)0)^*$
        \end{tabular} \\
    \begin{tabular}{@{} l@{}}
        $1^*11^*$
        \end{tabular} \CC & \begin{tabular}{@{} l@{}}
        The pattern consists of zero or more occurrences of the \\digit '1', followed by exactly one occurrence of the digit '1',\\ followed by zero or more occurrences of the digit '1'.
        \end{tabular} \CC &     \begin{tabular}{@{} l@{}}
        $1^*1^*1^*$
        \end{tabular}  \CC \\
    \begin{tabular}{@{} l@{}}
        $(1)^*0$
        \end{tabular} & \begin{tabular}{@{} l@{}}
        Zero or more occurrences of the character or group of \\characters before the asterisk.
        \end{tabular} &     \begin{tabular}{@{} l@{}}
        $(.^*)$
        \end{tabular} \\
        \bottomrule
    \end{tabular}
    \caption{Examples of errors the evaluated LLMs made while evaluating with \APPROACH{} for the regular expression dataset.}
    \label{tab:regex_error_examples}
\end{table}

\textbf{\FSNL{} Errors:} Most of the errors observed were the LLMs giving the wrong explanation, even for simple regular expressions. For example, GPT-4o and Claude Sonnet often described $c^*$ as "one or more occurrences of 'c'", where $c$ is a character from the alphabet (see example 1 in Table\,\ref{tab:regex_error_examples}). For the other LLMs, it was quite common for the explanation to not give the actual character (see example 4 in Table\,\ref{tab:regex_error_examples}). Overall, we observed a higher likelihood of SOTA LLMs hallucinating on regular expressions compared to the other datasets.

\textbf{\NLFS{} Errors:} The most common mistake when constructing a regular expression from natural language was misplacing $*$ or adding it when it was not needed (see example 3 in Table\,\ref{tab:regex_error_examples}). A mistake Claude Sonnet often made was adding a question mark to the end of the generated expression. Finally, even though we explicitly prompted the LLMs to use just $*$, sometimes the LLM would use $+$. 

\section{Standard Deviation Evaluation}
\label{appendix:std_dev_eval}

In this section, we perform an empirical analysis of the standard deviation of the syntactic compliance and accuracy of the results from using \APPROACH{}. Due to the 10 batches having different data, the standard deviation cannot be computed reliably based on the performance of the individual batches. 

\begin{figure}[h]
    \centering
    \includegraphics[width=\linewidth]{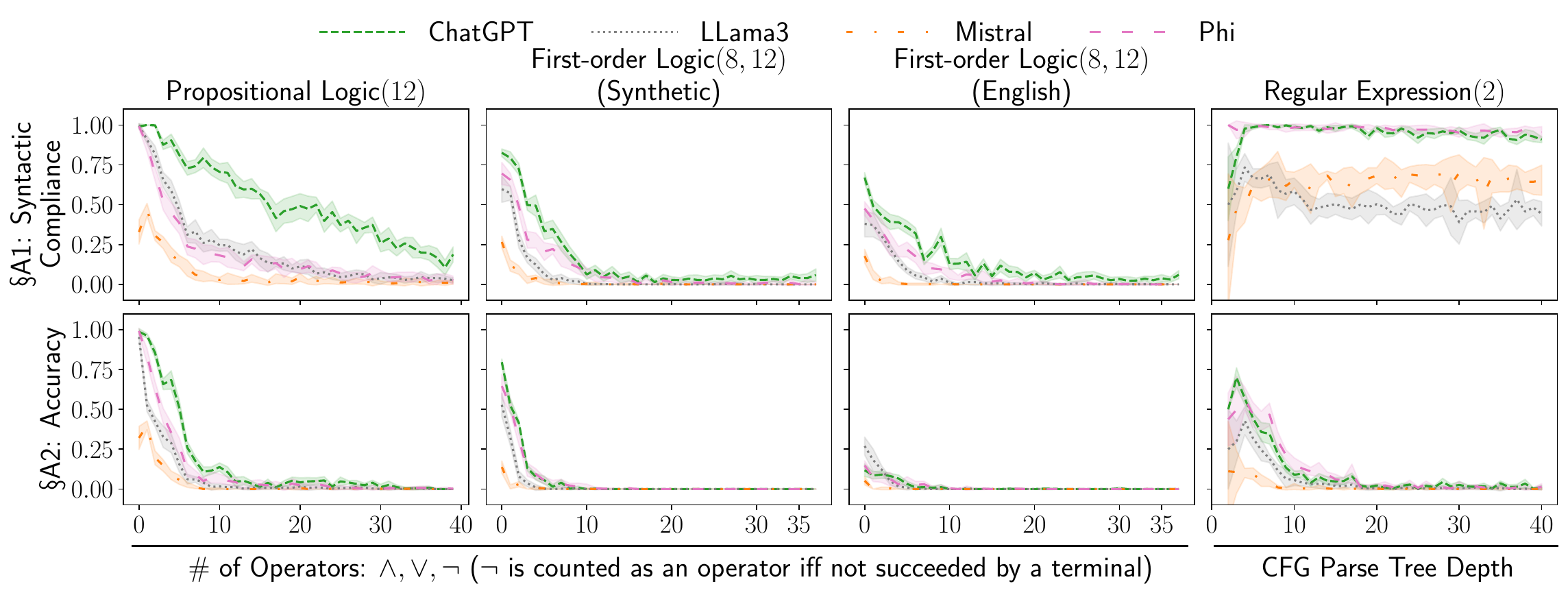}
        \vspace{-0.15in}
    \caption{Average and standard deviation error of Zero-shot Pass@1 results from using \APPROACH{} to assess LLMs w.r.t. \S A1, \S A2 (Sec.\,\ref{subsec:metrics}) on the first batch of the packaged datasets. The x-axis represents an increasing order of descriptional complexity.}
    \label{fig:dev_results}
\end{figure}

Due to limited cost and computational resources, we evaluate the standard deviation by running \APPROACH{} 10 times on the first batch of each dataset composed of 1974 propositional logic, 1900 first-order logic, and 1842 regular expressions examples. Additionally, we evaluated  GPT-3.5-turbo (ChatGPT) with a temperature of 1, LLama-3-8B-Instruct, Mistral-v0.2-7B-Instruct, and Phi-3-medium-4k-instruct. We calculated the mean and standard deviation of each independent run of \APPROACH{} and plotted the results in Figure\,\ref{fig:dev_results}.

For propositional and first-order logic, the standard deviation of the evaluated LLMs is not zero (due to non-deterministic text generation) but only caused noise of a couple of percentage points. While noisier, the standard deviation of the regular expression results was still less than 20\%. Overall, this experiment shows that the noise of non-deterministic text generation does not significantly impact evaluating with \APPROACH{} or the paper's results.

\section{Additional Zero-Shot Prompting Results}
\label{appendix:add_zero}

In this section, we evaluate other categorization metrics for the zero-shot prompting results. For the propositional and first-order logic datasets, the other metrics are the CFG parse tree depth needed to produce each FS expression and the individual number of each operator $(\land,\lor,\neg)$. For regular expressions, we have discussed in the main paper that each regular expression represents a minimal DFA that is unique up to isomorphism. Therefore, the other categorization metrics for regular expressions are the number of nodes $V$, the number of edges $E$, and the density of this minimal DFA. The density is calculated using Equation\,\ref{eq:dfa_density} where we discretize the value by rounding to every tenth.

\begin{equation}
    Density = \frac{|E|}{|V|(|V|-1)}
    \label{eq:dfa_density}
\end{equation}

\begin{figure}[h]
    \centering
    \includegraphics[width=0.75\linewidth]{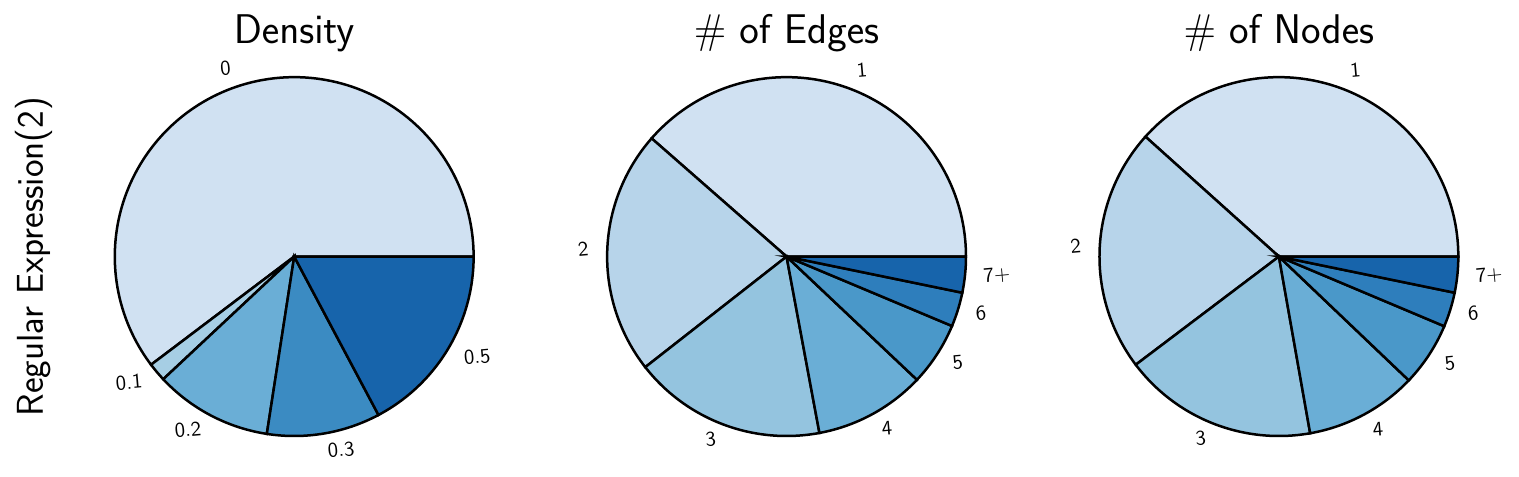}
    \caption{Count of the number of examples for each metric value for the regular expression datasets. The pie charts increase in values counter-clockwise while going from lighter to darker.}
    \label{fig:regex_dataset_bal}
\end{figure}

\begin{figure}
    \centering
    \includegraphics[width=\linewidth]{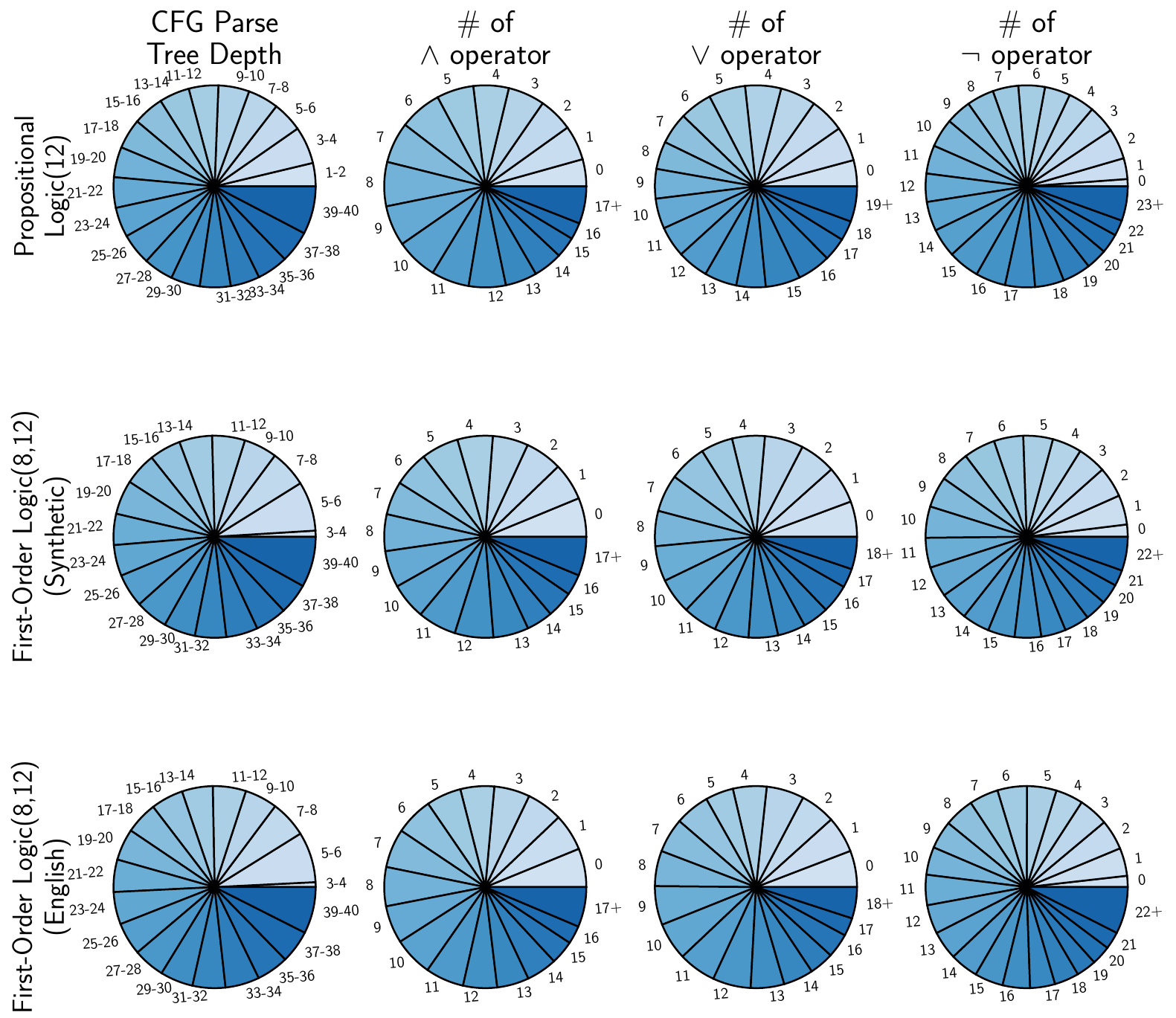}
    \caption{Count of the number of examples for each metric value for each of the datasets. Each row is a dataset and each column is a different metric that can be used to categorize the dataset. The pie charts increase in value counter-clockwise while going from lighter to darker.}
    \label{fig:dataset_bal_results}
\end{figure}

\textbf{Imbalanced Dataset Labels} Due to the datasets being created by sampling an equal number of expressions for each number of operators, taking this dataset and evaluating it in terms of the other metrics results in an imbalanced dataset. To examine this effect, we have created Figures\,\ref{fig:dataset_bal_results} and \ref{fig:regex_dataset_bal} to perform an analysis of how imbalanced the datasets are on these other metrics.

For the propositional and first-order logic other categorization metrics, the dataset is actually quite balanced due to CFG tree depth and the number of each individual operator having a high correlation to the total number of operators. As such, other than metric values close to the extrema, the noise from the imbalanced data will be marginal.

The regular expression dataset is less balanced due to a less direct correlation with the CFG tree depth. The middle of the density graphs will be the most noisy since there is significantly less data for densities of 0.1 and 0.2. The number of examples drops as the number of edges and nodes increases with less than 10\% of the data having more than 7 edges and/or nodes.

\begin{figure}
    \centering
    \includegraphics[width=\linewidth]{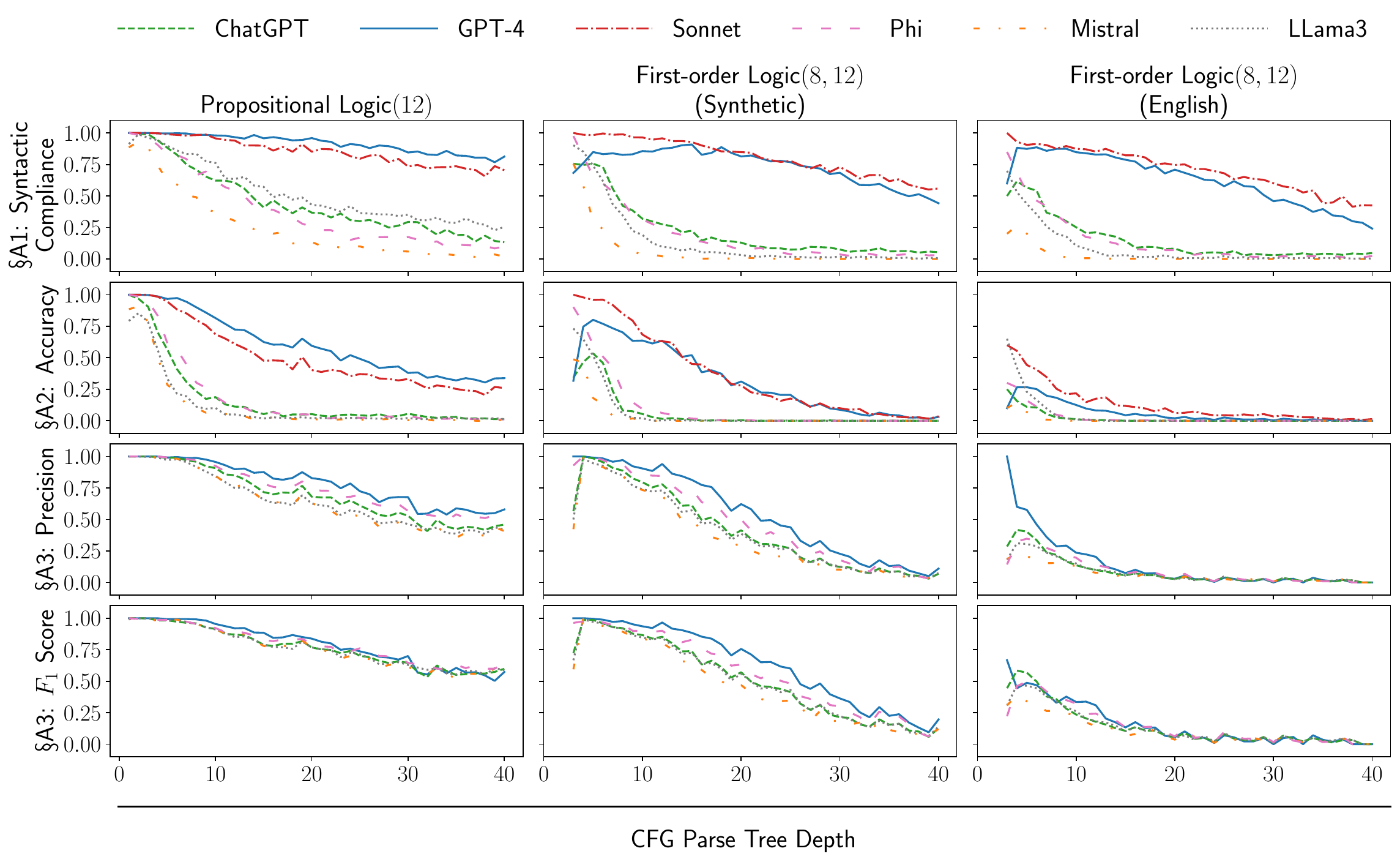}
    \caption{Zero-shot Pass@1 results (avg. over 10 batches, higher values better) from using \APPROACH{} to assess LLMs w.r.t. \S A1, \S A2, \S A3 (Sec.\,\ref{subsec:metrics}) on the packaged datasets. The x-axis is the depth of the CFG tree to produce the formula.}
    \label{fig:depth_zero_shot_results}
\end{figure}

\begin{figure}
    \centering
    \includegraphics[width=\linewidth]{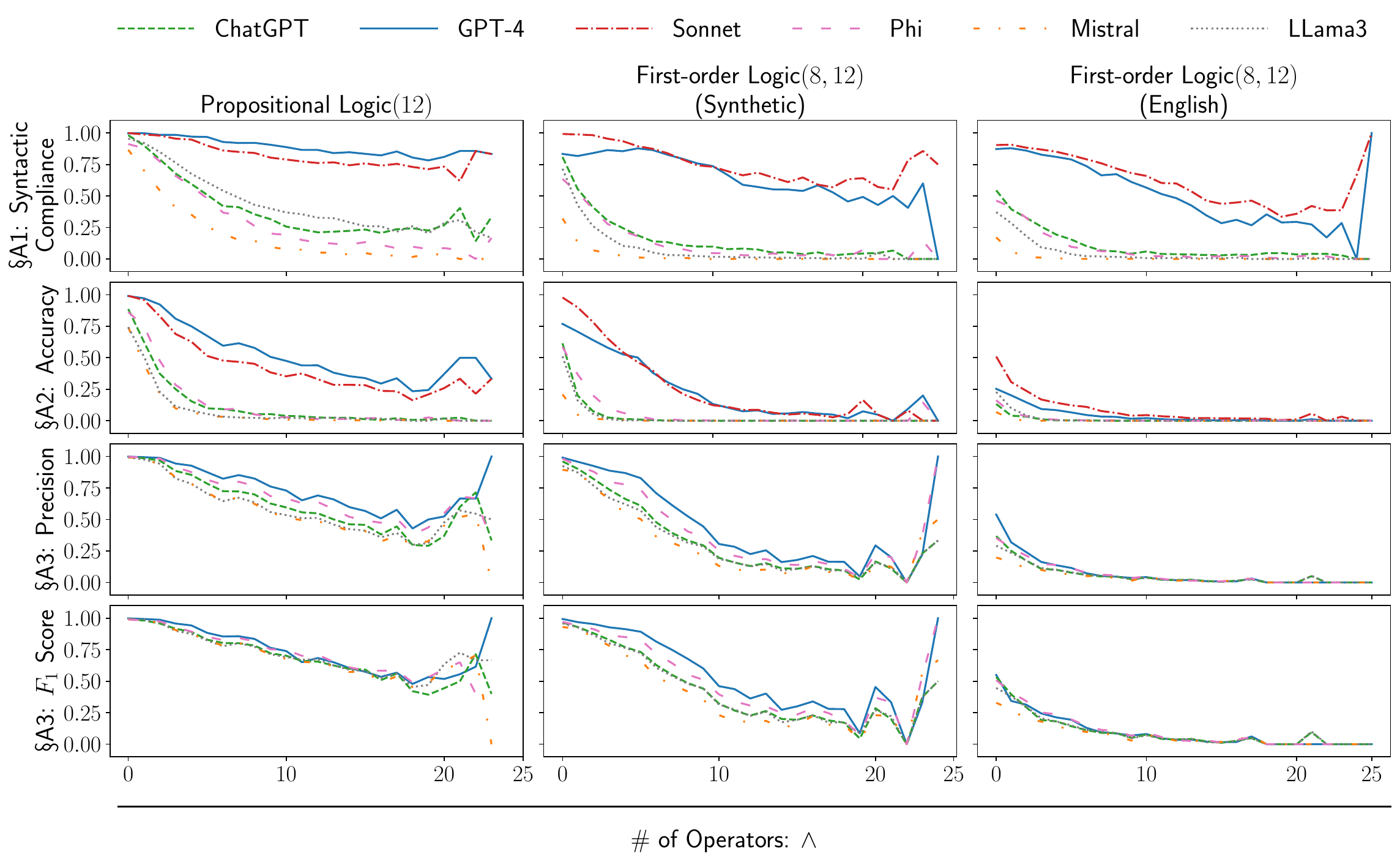}
    \caption{Zero-shot Pass@1 results (avg. over 10 batches, higher values better) from using \APPROACH{} to assess LLMs w.r.t. \S A1, \S A2, \S A3 (Sec.\,\ref{subsec:metrics}) on the packaged datasets. The x-axis is the number of and operators ($\land$) in the expression.}
    \label{fig:and_zero_shot_results}
\end{figure}

\begin{figure}
    \centering
    \includegraphics[width=\linewidth]{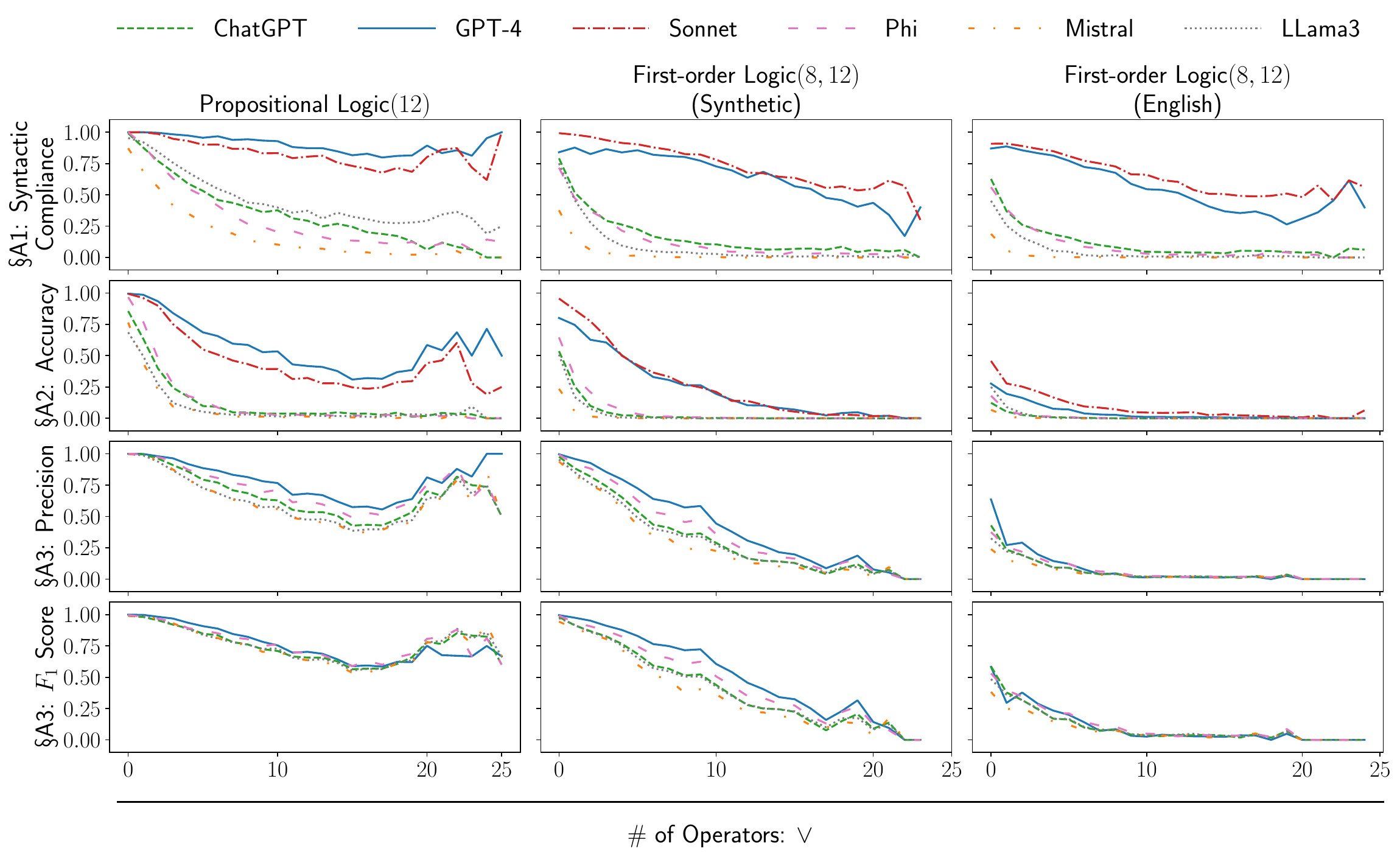}
    \caption{Zero-shot Pass@1 results (avg. over 10 batches, higher values better) from using \APPROACH{} to assess LLMs w.r.t. \S A1, \S A2, \S A3 (Sec.\,\ref{subsec:metrics}) on the packaged datasets. The x-axis is the number of or operators ($\lor$) in the expression.}
    \label{fig:or_zero_shot_results}
\end{figure}

\begin{figure}
    \centering
    \includegraphics[width=\linewidth]{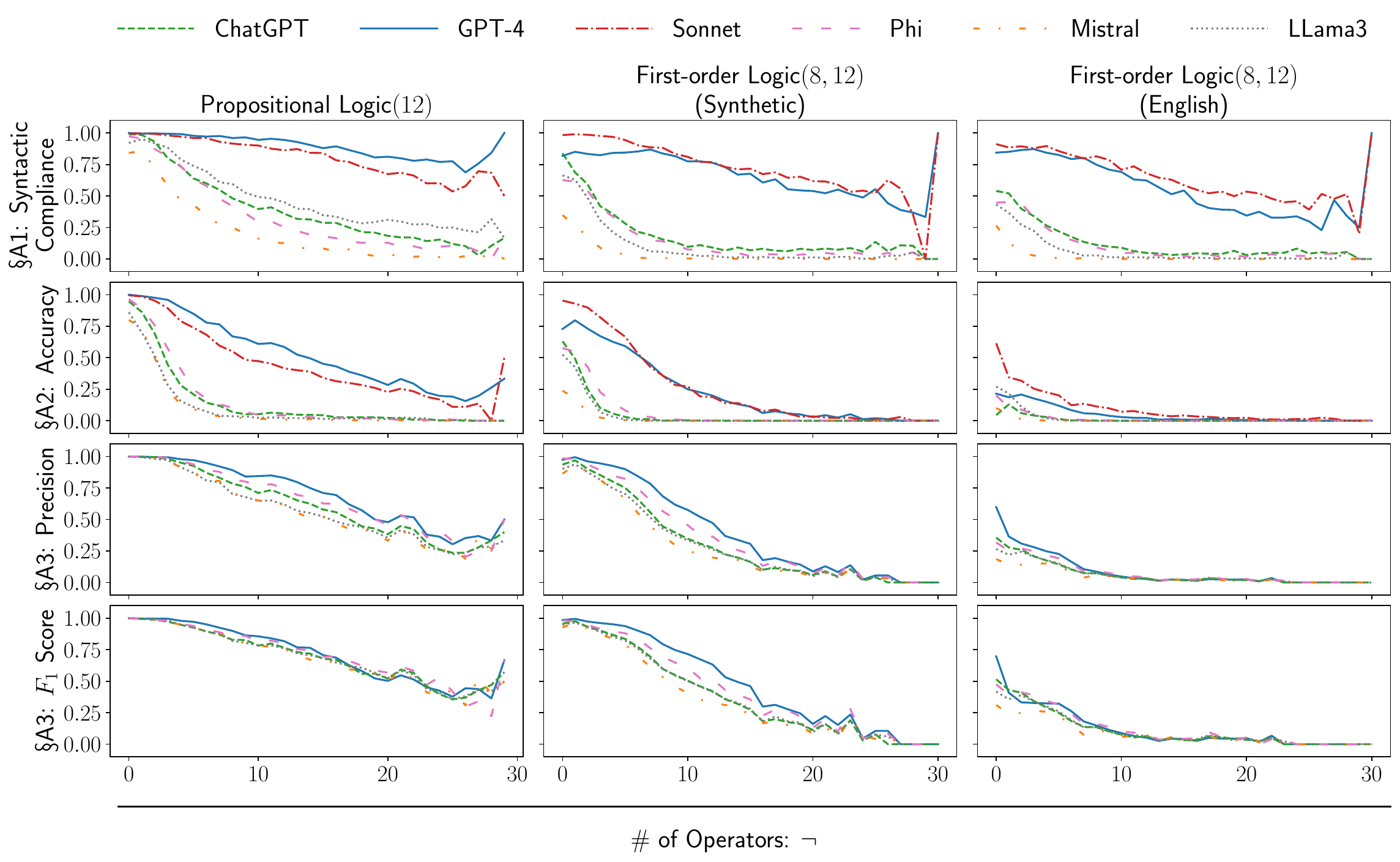}
    \caption{Zero-shot Pass@1 results (avg. over 10 batches, higher values better) from using \APPROACH{} to assess LLMs w.r.t. \S A1, \S A2, \S A3 (Sec.\,\ref{subsec:metrics}) on the packaged datasets. The x-axis is the number of negation operators ($\neg$) in the expression.}
    \label{fig:not_zero_shot_results}
\end{figure}

\begin{figure}
    \centering
    \includegraphics[width=\linewidth]{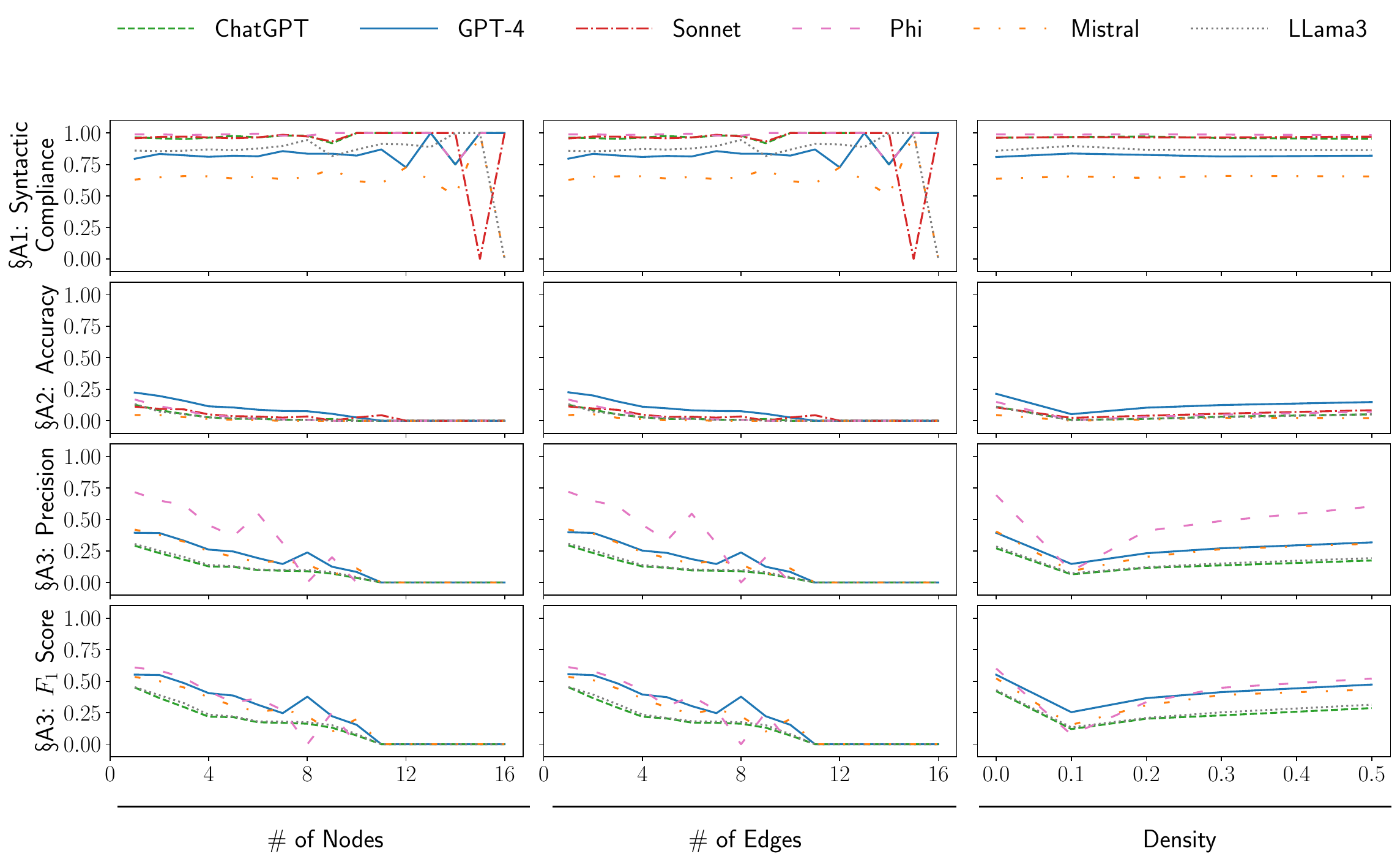}
    \caption{Zero-shot Pass@1 results (avg. over 10 batches, higher values better) from using \APPROACH{} to assess LLMs w.r.t. \S A1, \S A2, \S A3 (Sec.\,\ref{subsec:metrics}) on the packaged datasets. The x-axis is the depth of the CFG tree to produce the formula.}
    \vspace{-0.15in}
    \label{fig:regex_zero_shot_results}
    \vspace{-0.05in}
\end{figure}

\textbf{Categorization Metrics Performance} In Figures\,\ref{fig:depth_zero_shot_results}, \ref{fig:and_zero_shot_results},\ref{fig:or_zero_shot_results}, \ref{fig:not_zero_shot_results}, and \ref{fig:regex_zero_shot_results} the performance of each LLM over these other categorization metrics are shown. Across the board, we observe a diminishing performance regardless of the source of increasing complexity. Ignoring the noise from the low number of examples closer to the extrema, the depth of the tree showed a similar behavior as the operator number. Propositional logic performance was concave w.r.t the number of $\land$ and $\lor$ operators since it becomes easier to describe expressions composed of exclusively $\land$ and $\lor$ operators. However, this does not make it easier to remember all the grounded predicates, as shown by the continued decrease in performance in first-order logic. The negation operator was not concave, showing how LLMs struggle to handle multiple negation operators.

For regular expressions, increasing the number of nodes and edges reduces accuracy and the ability to evaluate equality. Density does not seem to be a factor, as the dip at 0.1 can be associated with noise due to the lower number of examples. Overall, these three metrics are much weaker factors in how well the LLM performs compared to the tree depth.

\section{Few-Shot Prompting Results}
\label{appendix:few_shot}

\begin{figure}
    \centering
    \includegraphics[width=\linewidth]{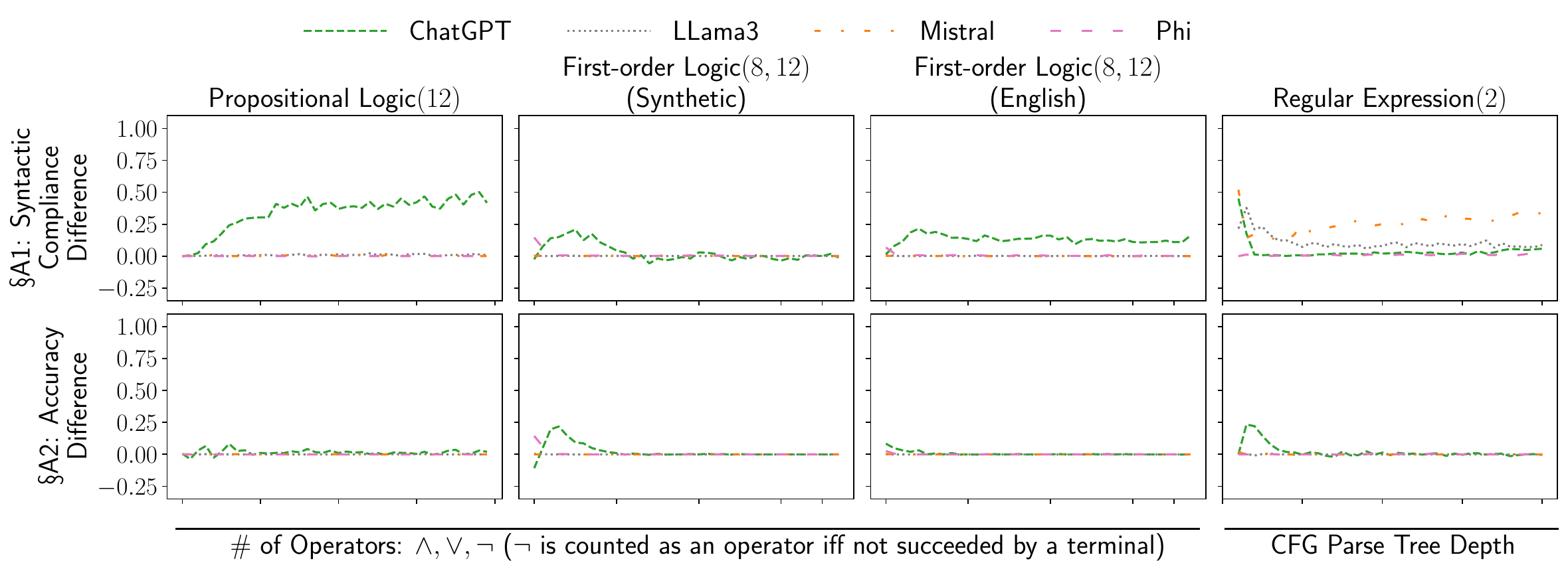}
    \vspace{-0.15in}
    \caption{Syntactic compliance and accuracy difference of few-shot Pass@1 compared to zero-shot Pass@1 results (avg. over 10 batches, higher values better) from using \APPROACH{} to assess LLMs w.r.t. \S A1, \S A2, \S A3 (Sec.\,\ref{subsec:metrics}) on the packaged datasets. The x-axis represents the increasing order of descriptional complexity.}
    \label{fig:few_shot_dif_results}
    \vspace{-0.25in}
\end{figure}

In this section, we discuss our few-shot prompting experiment and analyze the performance difference between zero-shot and few-shot prompting on \S A1 and \S A2.

Due to computational cost, we evaluated GPT-3.5-turbo (ChatGPT), LLama-3-8B-Instruct, Mistral-v0.2-7B-Instruct, and Phi-3-medium-4k-instruct on the same five datasets but inserted two examples into the prompts. First-order logic and predicate logic used the same two examples, while regular expressions used their own two examples.
In Figure\,\ref{fig:few_shot_dif_results}, the performance difference of each LLM when using few-shot prompting instead of zero-shot is shown. Using few-shot prompting increases syntactic compliance as the model has access to the desired format for encoding and decoding. For expressions with lower complexity, this translates to a higher accuracy. However, as complexity increases, the accuracy difference between zero-shot and few-shot prompting is negligible due to having the correct format for parsing but still failing to interpret the formal syntax expression correctly and/or compiling the interpretation correctly.

\end{document}